\documentclass{article}

\PassOptionsToPackage{numbers, sort&compress}{natbib}

\usepackage[preprint]{neurips_2023}

\usepackage[utf8]{inputenc} %
\usepackage[T1]{fontenc}    %
\usepackage{url}            %
\usepackage{booktabs}       %
\usepackage{amsfonts}       %
\usepackage{nicefrac}       %
\usepackage{microtype}      %

\usepackage{graphicx}
\usepackage{booktabs}
\usepackage{multirow}
\usepackage{multicol}
\usepackage{indentfirst}
\usepackage{subcaption}
\usepackage{makecell}
\usepackage{mdframed}

\usepackage[ruled,vlined]{algorithm2e}
\usepackage[dvipsnames]{xcolor}
\definecolor{commentcolor}{RGB}{59,116,116}   %
\newcommand{\PyComment}[1]{\ttfamily\textcolor{commentcolor}{\# #1}}  %
\newcommand{\PyCode}[1]{\ttfamily\textcolor{black}{#1}} %

\newcolumntype{M}[1]{>{\centering\raggedright\arraybackslash}m{#1}}
\usepackage{pifont}   %
\newcommand{\cmark}{\ding{51}} %
\newcommand{\xmark}{\textcolor{red}{\ding{55}}} %

\usepackage{hyperref}

\newcommand{\appref}[1]{\hyperref[#1]{Appendix~\ref*{#1}}}

\usepackage{nicematrix}
\definecolor{mygray}{gray}{0.85}
\definecolor{softgray}{rgb}{0.9, 0.9, 0.9}
\definecolor{softblue}{rgb}{0.88, 0.92, 1.0}
\definecolor{softgreen}{rgb}{0.88, 1.0, 0.88}
\definecolor{softyellow}{rgb}{1.0, 1.0, 0.88}
\definecolor{softred}{rgb}{1.0, 0.88, 0.88}
\definecolor{softpink}{rgb}{1.0, 0.88, 0.94}

\usepackage{color-edits}

\addauthor{gn}{magenta}
\addauthor{zq}{magenta}

\title{Evaluating Text-to-Visual Generation with Image-to-Text Generation}

\author{%
  {\bf Zhiqiu Lin$^{1}$} \quad {\bf Deepak Pathak$^1$} \quad {\bf Baiqi Li$^1$} \quad {\bf Jiayao Li$^1$} 
  \\  {\bf Xide Xia$^2$} \quad {\bf Graham Neubig$^1$} \quad {\bf Pengchuan Zhang$^{2*}$} \quad {\bf Deva Ramanan$^{1*}$} \\ 
  $^1$Carnegie Mellon University \quad\quad $^2$Meta \\ \href{https://linzhiqiu.github.io/papers/vqascore}{Code and models are open-sourced on our website}
}

\begin{document}

\maketitle

\begin{abstract}
Despite significant progress in generative AI, comprehensive evaluation remains challenging because of the lack of effective metrics and standardized benchmarks. For instance, the widely-used CLIPScore measures the alignment between a (generated) image and text prompt, but it fails to produce reliable scores for complex prompts involving compositions of objects, attributes, and relations. One reason is that text encoders of CLIP can notoriously act as a ``bag of words'', conflating prompts such as ~{\tt "the horse is eating the grass"} with {\tt "the grass is eating the horse"}~\cite{aro, winoground, lin2024revisiting}. To address this, we introduce the {\bf VQAScore}, which uses a visual-question-answering (VQA) model to produce an alignment score by computing the probability of a {\tt "Yes"} answer to a simple {\tt "Does this figure show \{text\}?"} question. Though simpler than prior art, VQAScore computed with off-the-shelf models produces state-of-the-art results across many (8) image-text alignment benchmarks. We also compute VQAScore with an in-house model that follows best practices in the literature. For example, we use a bidirectional image-question encoder that allows image embeddings to depend on the question being asked (and vice versa).  Our in-house model, {\bf CLIP-FlanT5}, outperforms even the strongest baselines that make use of the proprietary GPT-4V. Interestingly, although we train with only images, VQAScore can also align text with video and 3D models. VQAScore allows researchers to benchmark text-to-visual generation using complex texts that capture the compositional structure of real-world prompts. Towards this end, we introduce {\bf GenAI-Bench}, a more challenging benchmark with 1,600 compositional text prompts that require parsing scenes, objects, attributes, relationships, and high-order reasoning such as comparison and logic. GenAI-Bench also collects over 15,000 human ratings for leading image and video generation models such as Stable Diffusion, DALL-E 3, Midjourney, and Gen2. We open-source our data, model, and code at \href{https://github.com/linzhiqiu/t2v_metrics}{link}.
\end{abstract}

\begin{figure*}[ht!]
\centering
    \scalebox{0.95}{
        \begin{tabular}{c@{\hspace{8pt}}|@{\hspace{8pt}}c}
           (a) Computing VQAScore  &  (b) Types of image-question encoder  \\
           \includegraphics[width=0.46\textwidth]{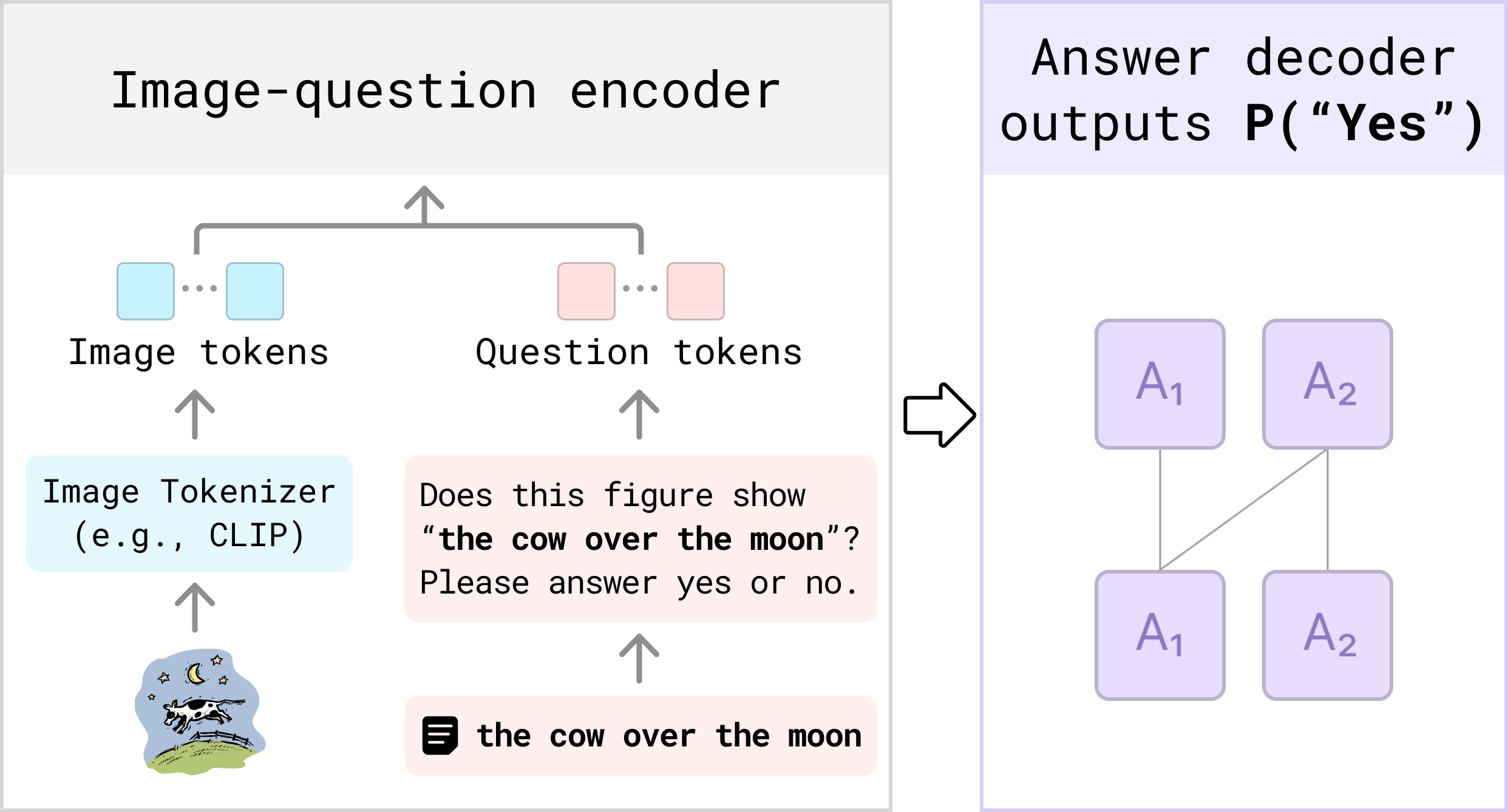} & \includegraphics[width=0.46\textwidth]{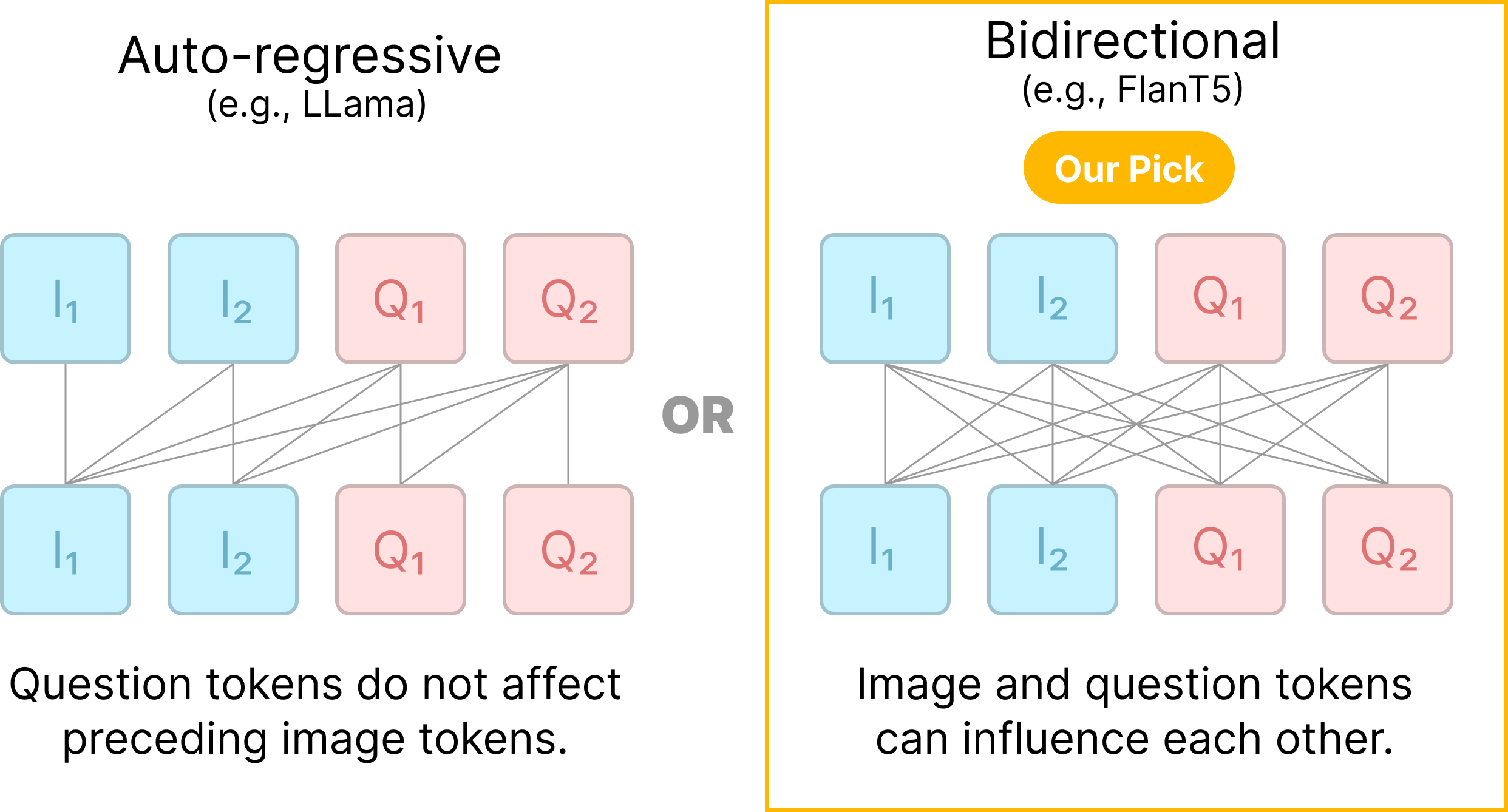}
        \end{tabular}
    }
    \caption{\small {\bf VQAScore.}  {\bf Figure  (a)} computes the VQAScore between an image and text by first converting the text into the question ``Does this figure show `\{{\tt text}\}'? Please answer yes or no.'' The image and question (after tokenization) are then fed into an image-question encoder, followed by an answer decoder that outputs the probability of ``Yes''. \appref{sec:vqascore_details} details the implementation and pseudocode. Our simple VQAScore based on off-the-shelf VQA models~\cite{llava15, instructblip} even rivals prior art that uses proprietary models~\cite{vq2, t2vscore, viescore} such as GPT4-Vision. {\bf Figure (b)} highlights the architectural choice of the image-question encoder. While popular open-source VQA models such as LLaVA-1.5~\cite{llava15} are derived from auto-regressive architectures like LLama-2~\cite{touvron2023llama} where question tokens do not affect preceding image tokens, we find it beneficial to adopt bidirectional encoders, e.g., FlanT5~\cite{flant5}. This allows the image to be ``looked at'' differently depending on the question, and vice versa. VQAScore based on our {\bf CLIP-FlanT5} model achieves a new state-of-the-art across text-to-image/video/3D alignment benchmarks. \autoref{fig:vqascore_teaser_1} shows examples of VQAScore's superior agreement with human judgments of images generated from complex text prompts.}
    \label{fig:vqascore_method}
\end{figure*}

\begin{figure*}[ht!]
\centering
    \begin{tabular}{c@{\hspace{2pt}}c}
       \includegraphics[width=0.49\textwidth]{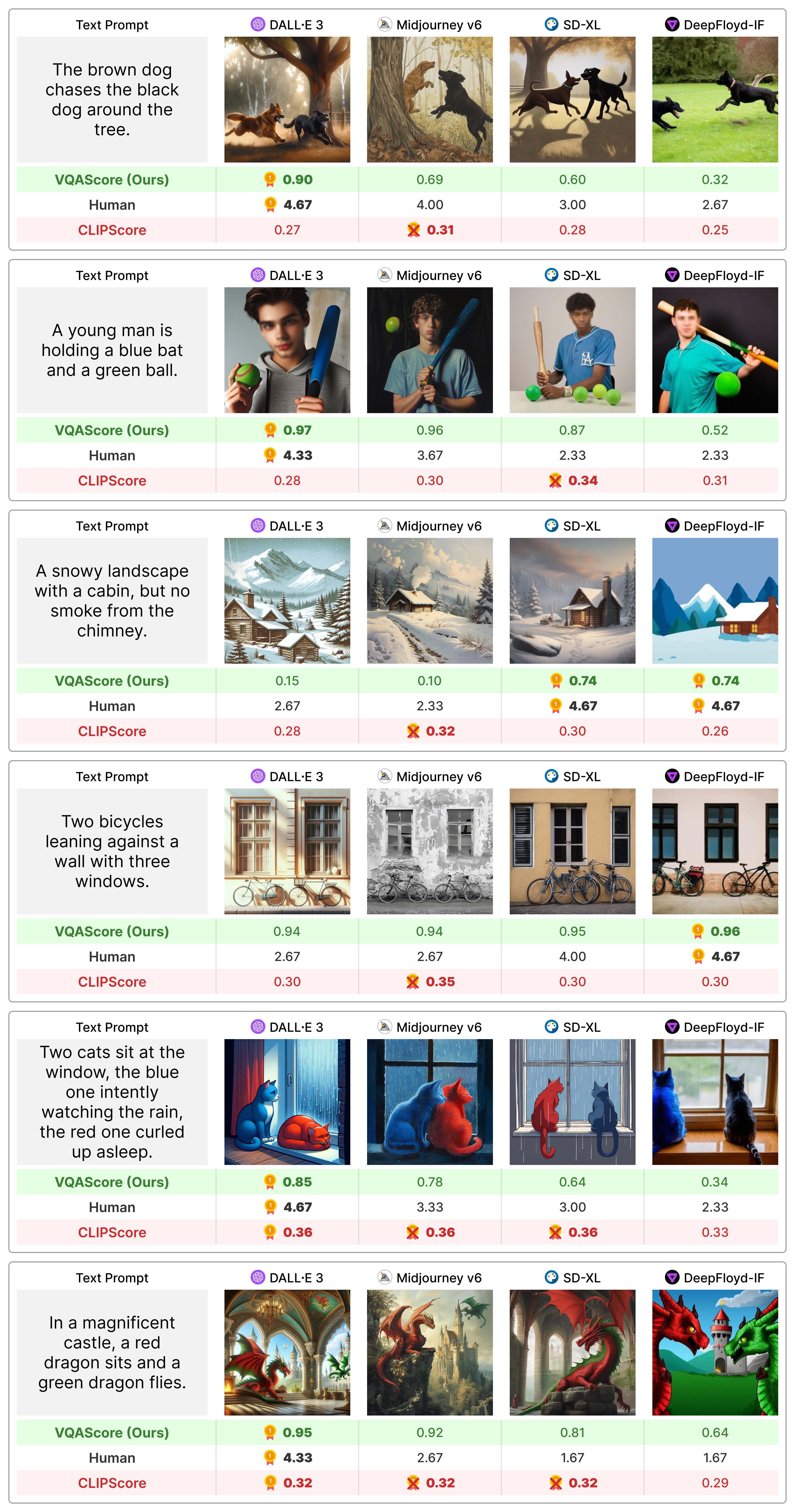}  &  \includegraphics[width=0.49\textwidth]{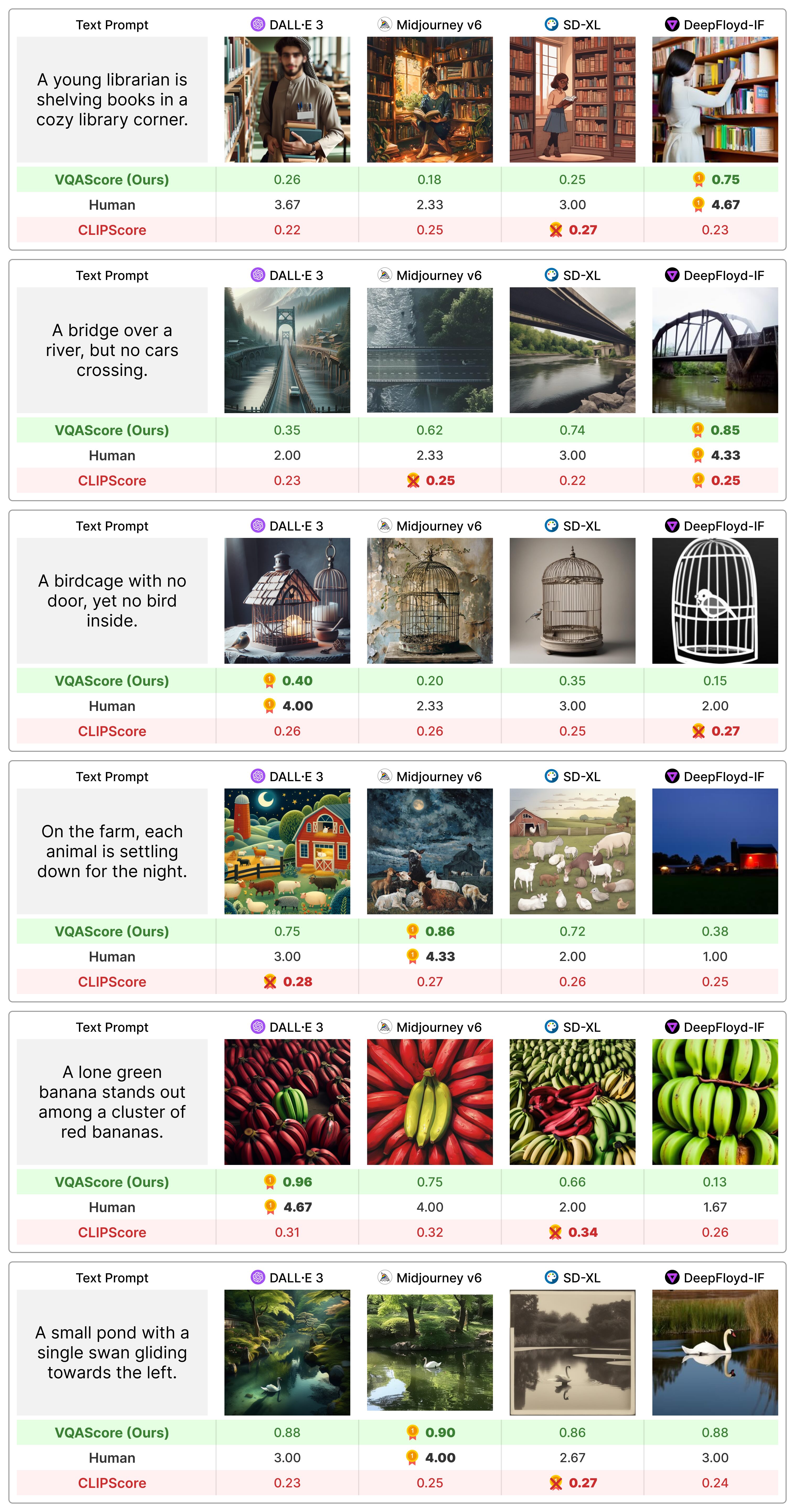} 
    \end{tabular}
    
    \caption{\small {\bf VQAScore (based on CLIP-FlanT5) versus CLIPScore} on samples from our GenAI-Bench (detailed in \autoref{sec:genai_bench}). GenAI-Bench consists of 1,600 text prompts spanning diverse compositional reasoning skills that challenge even leading models such as DALL-E 3~\cite{dalle3} and Stable Diffusion (SD)~\cite{stablediffusion}. VQAScore shows a significantly stronger agreement with human judgments compared to CLIPScore~\cite{clipscore}, making it a more reliable tool for automatic text-to-visual evaluation. We open-source our code and models for VQAScore at \href{https://github.com/linzhiqiu/t2v_metrics}{link}.
    }\vspace{-3mm}\label{fig:vqascore_teaser_1}
\end{figure*}

\section{Introduction}
\label{sec:intro}

Metrics play a key role in the evolution of science. For instance, perceptual metrics such as FID~\cite{FID}, IS~\cite{IS}, and LPIPS~\cite{LPIPS} have enabled tremendous progress by allowing researchers to systematically assess the {\em quality} of generated imagery. However, the generative AI community still lacks a robust metric that reveals how well an image {\em aligns} with an input text prompt. Indeed, generative models such as DALL-E 3~\cite{dalle3} and Gen2~\cite{gen2} produce remarkably photo-realistic images and videos that can still fail to align with input text prompts~\cite{tifa, pickscore, huang2023t2i}.

{\bf Challenges in evaluation.} Contemporary generative models~\cite{dalle3, imagen, controlnet, dai2023emu} primarily rely on {\em subjective} human evaluation~\cite{parti, imagen, singer2022make, liu2024language, yang2023idea2img} which can be expensive and difficult to reproduce. For systematic benchmarking, recent work~\cite{ruiz2023dreambooth, brooks2023instructpix2pix, chang2023muse, blipdiffusion, wu2023tune, singer2022make} adopts metrics such as CLIPScore~\cite{clip, clipscore}, which measures the (cosine) similarity of the embedded image and text prompt. However, accurately measuring vision-language alignment remains a significant challenge for even leading vision-language models (VLMs), because it requires advanced {\em compositional} reasoning skills (that may be as difficult as the underlying generative task!).
Studies~\cite{lin2024revisiting, aro, kamath2023text, ma2023crepe, eqben} show that VLMs like CLIP struggle with compositional text prompts involving multiple objects, attribute bindings, spatial/action relations, counting, and logical reasoning. Given the current state of the art, the power of standard evaluation metrics lags far behind the power of the generative models that they are evaluating.%

{\bf Decomposing texts via LLMs (prior art).} Recent neuro-symbolic methods~\cite{visprog, viper, vpeval, vq2, tifa, davidsonian} use off-the-shelf large language models (LLMs) like ChatGPT~\cite{gpt4, chatgpt} to tackle compositional reasoning through a {\em divide-and-conquer} approach, i.e., decomposing complex prompts into modular components. For example, visual programming~\cite{visprog, viper} uses LLMs to translate task instructions into symbolic programs, which themselves can invoke expert VLMs to return intermediate outputs like object counts~\cite{vpeval}. This inspires many recent methods~\cite{t2vscore, huang2023t2i, vpeval, vq2} to compute image-text alignment by decomposing the text prompt into simpler components, e.g., question-answer pairs. For example, TIFA~\cite{tifa} decomposes a prompt ``{\tt parent pointing at child}'' into questions like ``who is pointing at the child?'' and ``who is being pointed at?'', and returns the accuracy score of the answers generated by a visual-question-answering (VQA) model. However, these approaches struggle with more compositional text prompts, e.g., those from challenging benchmarks such as Winoground~\cite{winoground}. For example, given a prompt ``{\tt someone talks on the phone happily while another person sits angrily}'', the latest divide-and-conquer method Davidsonian~\cite{davidsonian} generates nonsensical questions like ``is the someone happy?'' and ``is there another person?''.

{\bf VQAScore (ours).}
Using recent VQA models based on multimodal LLMs~\cite{instructblip, llava}, we propose the following {\em end-to-end} approach that computes the generative likelihood~\cite{lin2024revisiting} of an answer to a simple question (see \autoref{fig:vqascore_method}). Given an {\tt image} and {\tt text}, we define their alignment to be the following probability: 
\begin{equation}
    P(\text{\small ``Yes''} | \text{\tt image}, \text{\small ``Does this figure show `\{{\tt text}\}'? Please answer yes or no.''}) \label{eq:vqa}
\end{equation}
We term this approach {\bf VQAScore}. Despite its simplicity, VQAScore implemented via open-source VQA models~\cite{llava15, instructblip} outperforms nearly all prior art including CLIPScore~\cite{clipscore}, models trained with extensive human feedback~\cite{imagereward,hpsv2,pickscore}, and divide-and-conquer methods~\cite{vpeval,tifa, vq2,davidsonian}. VQAScore even competes with approaches that rely on proprietary models~\cite{t2vscore, viescore} like GPT4-Vision trained on much larger datasets. We evaluate across a comprehensive suite of alignment benchmarks including Winoground~\cite{winoground}, EqBen~\cite{eqben}, TIFA160~\cite{tifa}, Flickr8K~\cite{clipscore}, DrawBench~\cite{imagen}, EditBench~\cite{wang2023imagen}, COCO-T2I~\cite{coco}, and Pick-a-Pic~\cite{pickscore}. 
We analyze the performance of various open-source models with respect to the benchmarks, and propose innovations in both modeling and benchmarking below.

{\bf What makes VQAScore effective?} To isolate factors crucial for image-text alignment, we train in-house VQA models controlling for architectures, training data, and training recipes. Recall that VQA models need be trained on (image, question, answer) examples~\cite{llava15}. We first point out that image-text alignment requires models to expose answer likelihoods rather than simply generate answer tokens (as much past work does~\cite{tifa, davidsonian}). Another crucial architectural choice is the type of image-question encoder. Many popular VQA models (e.g., LLaVA~\cite{llava, llava15}) are derived from next-token autoregressive LLMs (e.g., Llama-2~\cite{touvron2023llama}) where question embeddings depend on previously-encoded image tokens, but {\em not} vice versa. These are often known as uni-directional ``decoder-only'' architectures. However, we find it beneficial to allow visual embeddings to be influenced by the question being asked (and vice versa). Indeed, there exists tremendous evidence from neuroscience that humans parse imagery differently depending on the prompted task (via top-down feedback~\cite{hochstein2002view}). We operationalize this via a bidirectional ``encoder-decoder'' language model, FlanT5~\cite{flant5}. Specifically, we combine a pre-trained CLIP vision-encoder with a pre-trained FlanT5, which encodes image and question embeddings bidirectionally but generates answers auto-regressively (see \autoref{fig:vqascore_method}). By finetuning on public VQA datasets~\cite{llava15}, our final {\bf CLIP-FlanT5} sets a new state-of-the-art across all benchmarks. Interestingly, even though we need only simple question-answers at inference time~\eqref{eq:vqa}, VQAScore likely benefits from FlanT5's strong reasoning ability, trained on more than 400 language datasets with challenging question-answer pairs~\cite{flant5}.

\begin{figure*}[h!]
\centering
    \includegraphics[width=0.99\textwidth]{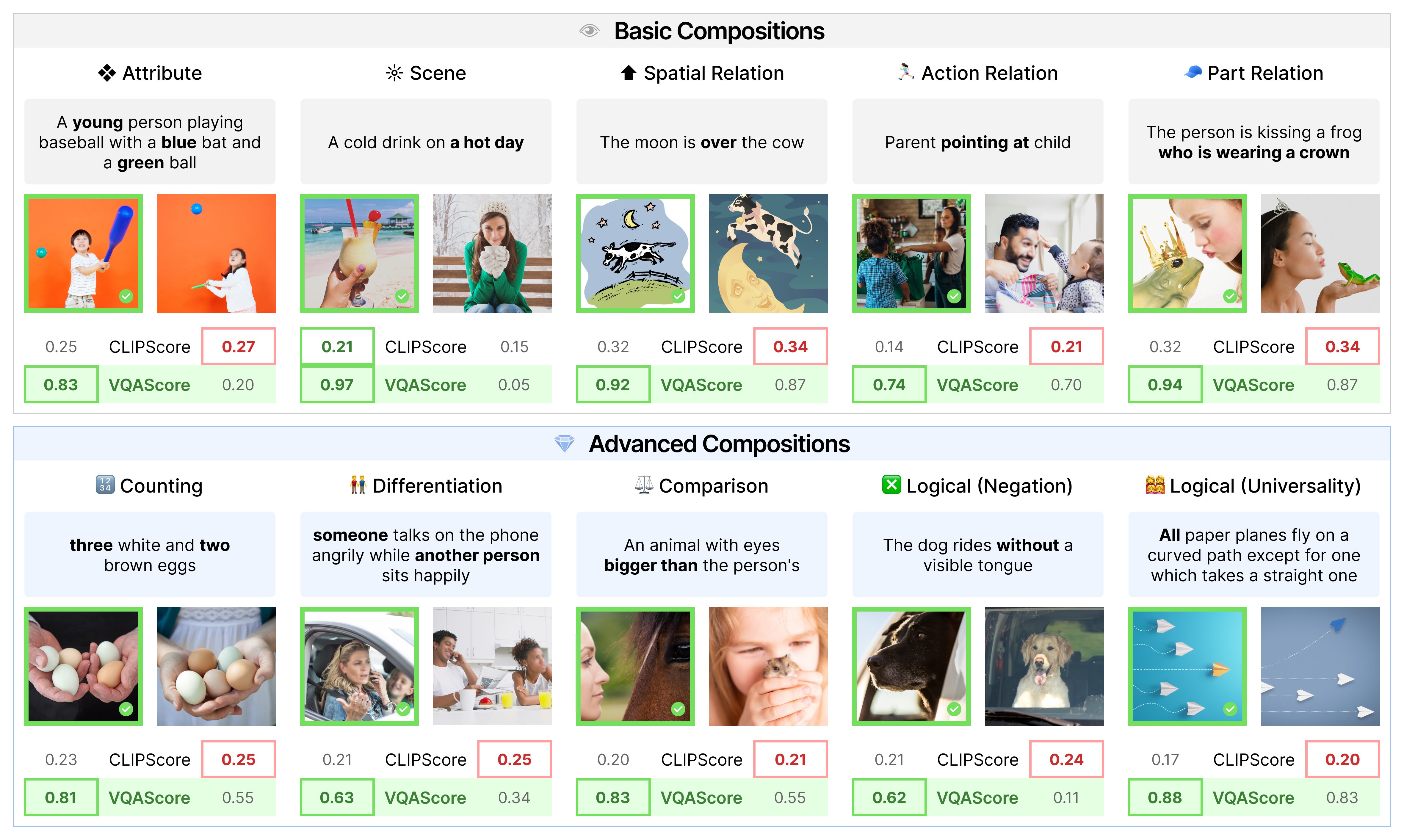}
    \caption{\small {\bf VQAScore (based on CLIP-FlanT5) versus CLIPScore} on random samples from the challenging Winoground~\cite{winoground} benchmark, containing real-world text prompts covering diverse compositional reasoning skills (which are carefully defined and labelled, as detailed in \appref{sec:skill_details}). VQAScore performs well across basic compositions (attribute/scene/relation) as well as advanced compositions that require higher-order reasoning, e.g., counting attribute-object pairs and reasoning logically over negation and universality statements. Quantitative performance per skill can be found in \autoref{tab:winoground_analysis}. 
    }\vspace{-3mm}\label{fig:vqascore_vs_baseline}
\end{figure*}

{\bf GenAI-Bench.}
We find that popular benchmarks for generative models~\cite{imagen, pickscore, huang2023t2i, t2vscore} like PartiPrompt~\cite{parti} do not capture the compositional structure of real-world text prompts (e.g., Winoground~\cite{winoground}). To remedy this, we identify a set of crucial skills for text-to-visual generation, ranging from basic (object, scene, attribute, and relation understanding) to advanced (comparison, differentiation, logical reasoning, and counting). \autoref{fig:vqascore_vs_baseline} presents illustrative examples~\footnote{Human faces are blurred to conceal identity.}. Although these skills frequently appear in user prompts, we find that existing benchmarks~\cite{tifa, parti, huang2023t2i} do not comprehensively cover them. To address the gaps, we introduce GenAI-Bench to evaluate both (1) text-to-visual generation models and (2) automated metrics. First, GenAI-Bench evaluates text-to-visual generation by collecting 1,600 prompts that cover essential visio-linguistic compositional reasoning skills. This allows us to identify the limitations of popular generative models such as Stable Diffusion, Midjourney, DALL-E 3, Pika, and Gen2. For quality purposes, the prompts are sourced from graphic designers who use text-to-visual tools in their profession. Next, GenAI-Bench evaluates automated metrics by collecting over 15,000 human ratings for ten leading text-to-visual models. GenAI-Bench exceeds the diversity and difficulty of prior benchmarks such as PartiPrompt~\cite{parti, pickscore, huang2023t2i}. We refer readers to \cite{ li2024evaluating} for further analysis on GenAI-Bench.

{\bf Extending to text-to-video/3D evaluation.} Finally, we conduct preliminary experiments on video-text and 3D-text alignment benchmarks~\cite{evalcrafter, wu2024gpt} by simply averaging the VQAScore across sampled frames or rendered views. VQAScore significantly surpasses popular methods such as CLIPScore~\cite{clipscore}, PickScore~\cite{pickscore}, and SOTA divide-and-conquer approaches that make use of GPT4-Vision~\cite{t2vscore}.

{\bf Contribution summary.}
\begin{enumerate}
    \item We propose {\bf VQAScore}, a simple metric that outperforms prior art without making use of expensive human feedback or proprietary models such as ChatGPT and GPT4-Vision.
    \item VQAScore based on our proposed 
    {\bf CLIP-FlanT5} model achieves the state-of-the-art in vision-language alignment, offering a strong alternative to CLIPScore. We open-source a pip-installable API at \href{https://github.com/linzhiqiu/t2i_metrics}{link} to run VQAScore for image/video/3D evaluation using one-line of Python code.
    \item We present {\bf GenAI-Bench}, a comprehensive benchmark with 1,600 compositional prompts to evaluate text-to-visual generation, surpassing the size and difficulty of existing benchmarks. Additionally, we provide over 15,000 human ratings (increasing to 80,000 in \cite{li2024evaluating}) to support research on vision-language alignment metrics. Our dataset is available at \href{https://huggingface.co/datasets/libaiqi/GenAI-Bench}{link}.
\end{enumerate}

\section{Related Works}

{\bf Automated text-to-visual evaluation.} Perceptual metrics like Inception Score (IS)~\cite{IS}, Fréchet Inception Distance (FID)~\cite{FID} and Learned Perceptual Image Patch Similarity (LPIPS)~\cite{LPIPS} use pre-trained networks to assess the quality of generated imagery. However, these metrics rely on reference images and do not generalize to vision-language alignment. Recent text-to-visual systems~\cite{imagen, kang2023scaling, chang2023muse, wu2023tune, singer2022make, ho2022imagen, dalle3, ruiz2023dreambooth, kawar2023imagic, gal2022image, kumari2023multi, meng2021sdedit, gal2022stylegan, brooks2023instructpix2pix, ku2023imagenhub, li2023seed, lee2023holistic} mostly report CLIPScore~\cite{clipscore}, which measures (cosine) similarity of the embedded image and text prompt. However, CLIP cannot reliably process compositional text prompts~\cite{winoground, aro, kamath2023text, lin2024revisiting}. Recent work further proposes three types of alignment metrics: {\bf (1) Human-feedback approach.} ImageReward~\cite{imagereward}, PickScore~\cite{pickscore}, and HPSv2~\cite{hpsv2} finetune VLMs like CLIP and BLIP on large-scale human ratings collected on generated images. {\bf (2) GPT4-Vision-based approach.} VIEScore~\cite{viescore} and GPT4-Eval~\cite{zhang2023gpt} carefully design a set of prompts for the proprietary GPT4-Vision~\cite{gpt4} to output an image-text alignment score. {\bf (3) Divide-and-conquer approach.} This popular line of methods~\cite{huang2023t2i, vpeval, llmscore, singh2023divide, t2vscore} use LLMs such as ChatGPT to decompose texts into simpler components for analysis. A notable technique within this framework is Question Generation and Answering (QG/A), exemplified by TIFA~\cite{tifa}, VQ2~\cite{vq2}, and Davidsonian~\cite{davidsonian}. For example, TIFA decomposes a text prompt into several simpler QA pairs and then outputs an alignment score as the accuracy of the answers generated by a VQA model.  

{\bf Visio-linguistic compositional reasoning.} Recent neuro-symbolic methods like visual programming~\cite{visprog, viper, hu2023visual} also use LLMs like ChatGPT to decompose complex visual tasks (described in natural language) into modular components. For instance, VPEval~\cite{vpeval} applies visual programming to compute image-text alignment, using ChatGPT to invoke expert VLMs like image captioning~\cite{blipv2} and open-vocabulary detection~\cite{groundingdino} models to examine fine-grained visual details. While visual programming achieves decent performance on classic benchmarks like GQA~\cite{gqa} and NLVR~\cite{nlvr}, we find that they rely heavily on hand-crafted in-context prompts (e.g., exemplar programs) and struggle on more challenging compositional tasks like Winoground~\cite{winoground}. Lastly, our VQAScore can be viewed as an extension of VisualGPTScore~\cite{lin2024revisiting}, which uses captioning models~\cite{blipv2} to calculate the generative likelihood of $P(\text{text}|\text{image})$.

\section{Image-Text Alignment Using VQAScore}
\label{sec:vqascore}
This section describes how we compute VQAScore for image-text alignment, and introduces our CLIP-FlanT5 model that achieves the state-of-the-art.

{\bf Image-text alignment.} Given an image {\bf i} and a text {\bf t}, we aim to compute an alignment score $S(\mathbf{i}, \mathbf{t}) \in \mathbb{R}$, where higher scores reflect greater image-text similarity. Ideally, a model-predicted alignment score should closely match human judgment. For example, given the text ``the moon is {\em over} the cow'', an image incorrectly showing the cow {\em above} the moon would likely receive a lower human rating.
\autoref{fig:vqascore_vs_baseline} provides such examples from the challenging Winoground~\cite{winoground} benchmark. However, this seemingly simple task challenges contrastive VLMs like CLIP~\cite{lin2024revisiting, aro, kamath2023text}, which fail to understand {\em compositional} text prompts involving relations, attributes, and logical reasoning. 
Instead, we propose using recent {\em generative} VLMs trained for visual-question-answering (VQA), which can reason compositionally by generating answers based on images and questions.

{\bf Computing VQAScore.} We calculate the alignment score directly from a VQA model using a simple template that converts the text {\bf t} to a question {\bf q}({\bf t}): 
\begin{align*}
\mathbf{t} &= \parbox[t]{.65\textwidth}{The moon is over the cow} \\
\mathbf{q}(\mathbf{t}) &= \parbox[t]{.65\textwidth}{Does this figure show "The moon is over the cow"? Please answer yes or no.} 
\end{align*}

\noindent 
Next, we compute the generative likelihood of ``Yes'' from the auto-regressive answer decoder of an off-the-shelf VQA model (see \autoref{fig:vqascore_method}-a):
\begin{align}
\text{VQAScore}(\mathbf{i}, \mathbf{t}) := P(``\text{Yes}"|\mathbf{i}, \mathbf{q}(\mathbf{t}))
\label{eq:vqascore_definition}
\end{align}

{\bf Improving VQAScore via CLIP-FlanT5.} While Eq.~\eqref{eq:vqascore_definition} can be readily computed using open-source models like LLaVA-1.5~\cite{llava15}, we improve VQAScore by training an in-house VQA model that follows best practices in the literature. Specifically, we find that popular VQA models~\cite{llava, llava15} are typically derived from ``decoder-only'' LLM architectures like Llama-2~\cite{touvron2023llama} that use a uni-directional (auto-regressive) attention mechanism, where each token is influenced only by its previous tokens, but not vice versa. However, literature in language modeling~\cite{flant5, ul2} suggests that bidirectional encoder-decoders (where all tokens can influence each other) outperform the uni-directional counterparts on reasoning-focused linguistic tasks~\cite{chia2023instructeval}. We argue that the architectural choice of image-question encoder becomes even more critical for visio-linguistic reasoning. For example, the state-of-the-art LLaVA-1.5~\cite{llava15} places image tokens (MLP-projected CLIP visual tokens) ahead of question tokens. This prevents question tokens from influencing the preceding image tokens, which contradicts how humans process visual information based on prompted tasks~\cite{hochstein2002view}. Although training a new bidirectional LLM from scratch is not feasible due to substantial computational costs, we can still improve VQAScore by replacing Llama-2 in LLaVA-1.5 with the state-of-the-art bidirectional encoder-decoder FlanT5~\cite{flant5} (see \autoref{fig:vqascore_method}-b for a comparison). For a fair comparison, we adhere to the training recipe of LLaVA-1.5, including the use of the same CLIP visual encoder, a modest 665K mixture of public VQA datasets, and a two-stage finetuning procedure. \appref{sec:clip_flant5_ablations} includes more training details.

\section{Experimental Results}
\label{sec:experiment_results}
This section outlines the experimental setup and results, highlighting VQAScore's superior performance compared to baseline methods such as CLIPScore~\cite{clipscore}, TIFA~\cite{tifa}, and PickScore~\cite{pickscore}.

{\bf Baseline methods.} We compare VQAScore against five popular method types: (1) VLM-based metrics (CLIPScore~\cite{clipscore} and BLIPv2Score~\cite{blipv2}); (2) VLMs finetuned on human feedback (PickScore~\cite{pickscore}, ImageReward~\cite{imagereward}, and HPSv2~\cite{hpsv2}); (3) visual programming methods (VisProg~\cite{visprog}, ViperGPT~\cite{viper}, and VPEval~\cite{vpeval}); (4) divide-and-conquer methods using VQA (TIFA~\cite{tifa}, VQ2~\cite{vq2}, and Davidsonian~\cite{davidsonian}); (5) approaches using proprietary models like GPT4-Vision (GPT4V-Eval~\cite{zhang2023gpt} and VIEScore~\cite{viescore}). \appref{sec:method_details} includes the implementation details of these methods.

{\bf Evaluating VQAScore on compositional image-text matching.}
We begin with the two most challenging benchmarks, Winoground~\cite{winoground} and EqBen~\cite{eqben}, where each test sample has two (image, text) pairs. These benchmarks evaluate image-text matching through binary retrieval tasks that identify the best caption (from the pair of candidates) for a given image, and vice versa. Importantly, the benchmark API requires algorithms to return a match score for each candidate (image, text) pair instead of a relative ranking. This means they can be readily repurposed to evaluate image-text alignment. Compared to existing alignment benchmarks~\cite{tifa}, we find that these matching benchmarks (especially Winoground) include more challenging text prompts with compositional structure (inspiring our own benchmarking efforts in \autoref{sec:genai_bench}).
For example, the prompt {\tt ``someone talks on the phone angrily while another person sits happily''} requires the model to differentiate between two people (entities) based on emotions (attributes) and actions (relations). Another prompt {\tt ``three white and two brown eggs''} requires the model to count attribute-object pairs. \autoref{fig:vqascore_vs_baseline} compares VQAScore and CLIPScore on random Winoground examples. \autoref{sec:skill_details} provides an in-depth analysis of the skills covered by these benchmarks.

\begin{table*}[h!]
\centering
\renewcommand{\arraystretch}{1.3}
\caption{\textbf{VQAScore achieves SOTA performance on challenging image-text matching benchmarks that require advanced compositional reasoning.} We thoroughly ablate our proposed VQAScore with popular recent approaches on Winoground~\cite{winoground} and EqBen~\cite{eqben}. We strictly adhere to the original evaluation protocols and report text/image/group scores, with higher scores indicating better performance. We describe these metrics in \appref{sec:eval_details}. Note that our VQAScore (highlighted in \textcolor{green}{green}) even matches or outperforms proprietary models (highlighted in \textcolor{gray}{gray}) that appear to be trained on much more data (such as PALI-17B~\cite{chen2022pali} and GPT4-Vision~\cite{gpt4}).}
\scalebox{0.8}{
\begin{NiceTabular}{llcccccccccc}
            \CodeBefore
              \rectanglecolor{softgray}{20-1}{25-8}
              \rectanglecolor{softgreen}{25-1}{29-8}
            \Body
\toprule[1.5pt]
\multirow{2}{*}{\textbf{Method}}  & \multirow{2}{*}{\textbf{Models}} & \multicolumn{3}{c}{\textbf{Winoground}} &
\multicolumn{3}{c}{\textbf{EqBen}}
\\ 
\cmidrule(l){3-5} \cmidrule(l){6-8}
  &   &      Text & Image & Group & Text & Image  & Group  \\ \midrule
\multicolumn{5}{l}{\textit{Baselines}} \\
Random Chance & -- & 25.0 & 25.0 & 16.7 & 25.0 & 25.0 & 16.7 \\
Human Evaluation & -- & 89.5 & 88.5 & 85.5 & -- & -- & -- \\ \midrule
\multicolumn{5}{l}{\textit{{Based on vision-language models}}} \\
\multirow{1}{*}{CLIPScore~\cite{clip}}  & CLIP-L-14 & 27.8 & 11.5 & 7.8 & 35.0 & 35.0 & 25.0 \\
 \multirow{1}{*}{BLIPv2Score~\cite{blipv2}} & BLIPv2 & 43.3 & 21.3 & 17.5 & 48.6 & 43.6 & 35.0 \\ \midrule
\multicolumn{5}{l}{\textit{Finetuned on human feedback}} \\
\multirow{1}{*}{PickScore~\cite{pickscore}} & CLIP-H-14 (finetuned) & 23.8 & 12.5 & 6.8 & 35.7 & 39.3 & 23.6 \\
\multirow{1}{*}{ImageReward~\cite{imagereward}} & BLIPv2 (finetuned) & 42.8 & 15.3 & 12.8 & 37.9 & 36.4 & 26.4 \\
\multirow{1}{*}{HPSv2~\cite{hpsv2}} & CLIP-H-14 (finetuned) & 11.5 & 7.8 & 4.0 & 27.9 & 26.4 & 17.1 \\ \midrule
\multicolumn{5}{l}{\textit{Based on visual programming}} \\
VisProg~\cite{visprog} & ChatGPT, ViLT, OWL-ViT & 3.5 & 3.5 & 3.5 & 7.9 & 7.9 & 7.9 \\
ViperGPT~\cite{viper} & ChatGPT, CLIP, BLIP, GLIP & 7.8 & 7.8 & 7.8 & 4.3 & 4.3 & 4.3 \\
VPEval~\cite{vpeval} & ChatGPT, BLIP, GroundDINO & 12.8 &	11.0 &	6.3 &	34.3 &	25.7 &	21.4  \\\midrule
\multicolumn{4}{l}{\textit{Divide-and-conquer via VQA}} \\
\multirow{1}{*}{VQ2~\cite{vq2}} & FlanT5, LLaVA-1.5 & 14.0 &	27.3 &	10.0 &	22.9 &	40.7 &	20.0 \\
\multirow{1}{*}{Davidsonian~\cite{davidsonian}} & ChatGPT, LLaVA-1.5 & 21.0 & 16.8 & 15.5 & 26.4 & 20.0 & 20.0 \\ \midrule
\multicolumn{5}{l}{\textit{Based on proprietary models}} \\ 
\multirow{1}{*}{TIFA~\cite{tifa, vq2}} & Llama-2, PaLI-17B &  19.0 & 12.5 & 11.3 & -- & -- & -- \\
VQ2~\cite{vq2} & FlanT5, PaLI-17B & 47.0 & 42.0 & 30.5 & -- & -- & -- \\
GPT4V-Eval~\cite{zhang2023gpt} & GPT4-Vision & 44.5 &	49.0 & 36.3 & 42.9 & 40.0 &	35.0 \\
VIEScore~\cite{viescore} & GPT4-Vision & 40.8 &	39.3 & 34.5 & 40.0 & 34.3 &	32.9 \\\midrule
\multicolumn{4}{l}{\textit{VQAScore (ours) using open-source VQA model}} \\
\multirow{1}{*}{\bf VQAScore} & InstructBLIP & 44.5 &	42.8 &	28.5 &	49.3 &	58.6 &	38.6 \\ %
\multirow{1}{*}{\bf VQAScore} & LLaVA-1.5 & 45.5 &	41.3 &	29.8 &	42.9 &	60.0 &	35.0 \\ \midrule
\multicolumn{4}{l}{\textit{VQAScore (ours) using our VQA model}} \\
\multirow{1}{*}{\bf VQAScore} & {\bf CLIP-FlanT5 (Ours)} & {\bf 60.0} &	{\bf 57.5} &	{\bf 46.0} &	{\bf 59.3} &	{\bf 63.6} &	{\bf 47.9} \\

\bottomrule[1.5pt]
\end{NiceTabular}
}
\label{tab:vqascore_results_winoground_eqben}
\end{table*}

{\bf VQA achieves SOTA on matching benchmarks.} \autoref{tab:vqascore_results_winoground_eqben} shows that VQAScore sets a new state-of-the-art on both benchmarks. Compared to baselines (e.g., CLIPScore~\cite{clipscore} and PickScore~\cite{pickscore}) that perform at chance-level, our VQAScore achieves 2x to 5x higher scores. Our results using open-source VQA models (e.g., InstructBLIP~\cite{instructblip} and LLaVA-1.5~\cite{llava15}) can match the previous SOTA method VQ2~\cite{vq2} that uses the closed-source PaLI-17B~\cite{chen2022pali} model, which was trained on 40x more private data (over 20 billion images and texts). Crucially, VQAScore based on our in-house CLIP-FlanT5 model surpasses all prior art, including two recent methods~\cite{zhang2023gpt, viescore} that use the proprietary (and expensive) GPT4-Vision~\cite{gpt4} to score image-text alignment
. Moreover, our experiments show that visual programming methods, including VisProg~\cite{visprog}, ViperGPT~\cite{viper}, and VPEval~\cite{vpeval}, fail at compositional image-text matching, despite utilizing ChatGPT with expert VLMs~\cite{blipv2, groundingdino}. To fairly compare with divide-and-conquer methods that also use VQA models, we evaluate VQAScore against them based on the same VQA architectures, as demonstrated below.

\begin{table*}[h]
\centering
\renewcommand{\arraystretch}{1.3}
\caption{\textbf{Comparing VQAScore against divide-and-conquer methods using the same VQA models.} For a fair comparison, we apply both VQAScore and three open-source divide-and-conquer methods (TIFA~\cite{tifa}, VQ2~\cite{vq2}, and Davidsonian~\cite{davidsonian}) to the same underlying VQA architectures (InstructBLIP, LLaVA-1.5, and our CLIP-FlanT5).
These popular methods make use of large language models to decompose compositional text prompts into simpler question-answer pairs for analysis, 
e.g., Llama-2 for TIFA, FlanT5 for VQ2, and ChatGPT for Davidsonian. However, we find that they still struggle on compositional text prompts and often generate nonsensical question-answer pairs (more analysis can be found in \appref{sec:method_details}). Our end-to-end VQAScore (highlighted in \textcolor{green}{green}) outperforms them all using a much simpler question-answer template.  %
}
\scalebox{0.8}{
\begin{NiceTabular}{llcccccccccc}
            \CodeBefore
              \rectanglecolor{softgreen}{7-2}{7-8}
              \rectanglecolor{softgreen}{11-2}{11-8}
              \rectanglecolor{softgreen}{15-2}{15-8}
            \Body
\toprule[1.5pt]
\multirow{2}{*}{\textbf{VQA Model}} & 
\multirow{2}{*}{\textbf{Method}} & \multicolumn{3}{c}{\textbf{Winoground}} &
\multicolumn{3}{c}{\textbf{EqBen}}
\\ 
\cmidrule(l){3-5} \cmidrule(l){6-8}
  &   &    Text & Image & Group & Text & Image  & Group  \\ \midrule
\multirow{1}{*}{--} & Random Chance & 25.0 & 25.0 & 16.7 & 25.0 & 25.0 & 16.7 \\ \midrule
\multirow{4}{*}{InstructBLIP-FlanT5-11B~\cite{instructblip}} & \multirow{1}{*}{TIFA~\cite{tifa}}  & 20.3 &	16.3 &	14.5 &	25.0 &     25.7 &     18.6 \\
& \multirow{1}{*}{VQ2~\cite{vq2}}  & 19.0 &	26.3 &	11.3 &	20.0 &     39.3 &     15.7 \\
& \multirow{1}{*}{Davidsonian~\cite{davidsonian}} &18.3 &	15.3 &	14.0 &	22.1 &     17.9 &     15.7 \\
& {\bf VQAScore (ours)} & {\bf 44.5} &	{\bf 42.8} & {\bf 28.5} & {\bf 49.3} &	{\bf 58.6} & {\bf 38.6}\\ \midrule
\multirow{4}{*}{LLaVA-1.5-13B~\cite{llava}} & \multirow{1}{*}{TIFA~\cite{tifa}}  & 22.8 &	18.5 &	15.5 &	30.0 &	30.0 &	21.4 \\
& \multirow{1}{*}{VQ2~\cite{vq2}}  & 14.0 & 27.3 &	10.0 &	22.9 &	40.7 &	20.0 \\
& \multirow{1}{*}{Davidsonian~\cite{davidsonian}} &21.0 &	16.8 &	15.5 &	26.4 &	20.0 &	20.0 \\
& {\bf VQAScore (ours)} & {\bf 45.5} &	{\bf 41.3} & {\bf 29.8} & {\bf 42.9} &	{\bf 60.0} & {\bf 35.0}\\ \midrule
\multirow{4}{*}{CLIP-FlanT5-11B (Ours)} & \multirow{1}{*}{TIFA~\cite{tifa}}  & 26.5 & 19.3 & 16.0 & 28.6 & 23.6 & 18.6 \\
& \multirow{1}{*}{VQ2~\cite{vq2}}  & 19.8 & 30.3 & 14.0 & 25.7 & 47.1 & 22.1 \\
& \multirow{1}{*}{Davidsonian~\cite{davidsonian}} & 16.3 & 11.5 & 9.8 & 17.1 & 11.4 & 11.4 \\
& {\bf VQAScore (ours)} & {\bf 60.0} &	{\bf 57.5} &	{\bf 46.0} &	{\bf 59.3} &	{\bf 63.6} &	{\bf 47.9} \\
\bottomrule[1.5pt]
\end{NiceTabular}
}
\label{tab:vqa_based_methods_on_winoground_eqben}
\end{table*}

{\bf End-to-end VQAScore outperforms divide-and-conquer methods.} For a fair comparison, we apply three popular open-source divide-and-conquer methods (TIFA~\cite{tifa}, VQ2~\cite{vq2}, Davidsonian~\cite{davidsonian}) with the same VQA models used for VQAScore.
These methods either carefully prompt ChatGPT or finetune open-source LLMs like Llama-2 to decompose texts into simpler question-answer pairs. However, we discover that they struggle with compositional texts. For example, 
given {\tt ``someone talks on the phone angrily while another person sits happily''}, Davidsonian~\cite{davidsonian} asks nonsensical questions like ``is the someone talking angrily?'' and ``is the someone talking on the phone?''. Similarly, VQ2~\cite{vq2} asks silly questions like ``who talks with angrily on the phone?'' and expects an answer of ``someone''. Additionally, we find it crucial to expose the answer likelihood~\cite{lin2024revisiting, wu2023q}, which is less biased than generating multiple-choice answers as done by \cite{tifa, davidsonian}. For instance, LLaVA-1.5~\cite{llava15} biases towards answering ``Yes'' to $80\%$ of  the questions should be answered with ``No''  on Winoground (with the questions generated by Davidsonian~\cite{davidsonian}). \autoref{sec:method_details} presents more analysis of these methods. \autoref{tab:vqa_based_methods_on_winoground_eqben} confirms that our simpler VQAScore significantly outperforms the more complex divide-and-conquer methods regardless of the underlying VQA models.

{\bf VQAScore can more effectively handle compositional text prompts.} For a detailed analysis, we tag each Winoground sample by its associated compositional reasoning skills. \autoref{sec:skill_details} describes the labeling policy and procedure. \autoref{tab:winoground_analysis} shows that VQAScore based on our CLIP-FlanT5 model significantly surpasses CLIPScore by 5x in basic skills (e.g., attribute, scene, relation) and 10x in advanced skills (e.g., counting, comparison, differentiation, negation, universality). Though trained on the same VQA data, our CLIP-FlanT5 (based on the 11B FlanT5 model) consistently outperforms LLaVA-1.5 (based on the 13B Llama-2 model). We believe our model benefits from the bidirectional image-question encoding and strong language capabilities of FlanT5, which has been finetuned on over 400 complex QA datasets~\cite{flant5}. \appref{sec:clip_flant5_ablations} further demonstrates that VQAScore can be improved by scaling up the language model and finetuning on more VQA data.

\begin{table*}[h!]
\centering
\renewcommand{\arraystretch}{1.3}
\caption{{\bf Fine-grained analysis on Winoground.} We report group scores per skill category. Note that each sample can naturally incorporate multiple skills. For instance, ``{\tt a white dog is on a brown couch}'' involves understanding both ``attribute'' and ``spatial relation''. Additionally, a more complex prompt like ``{\tt six people wear blue shirts and no people wear white shirts}'' requires higher-order reasoning (e.g., ``counting'' and ``negation'') along with other basic skills. We detail the skill definitions in \appref{sec:skill_details}. Notably, the ``advanced'' skills (e.g., logic and comparison) prove more difficult (indicated by lower overall scores) compared to the ``basic'' skills. Our CLIP-FlanT5-based VQAScore excels across all skills -- 5x better than CLIPScore on ``basic skills'' and 10x better on ``advanced skills''.
}
\scalebox{0.51}{
\begin{tabular}{cc}
    \scalebox{1.0}{
    \begin{NiceTabular}{lccccc|c}
        \CodeBefore
          \rectanglecolor{softgreen}{4-1}{7-7}
        \Body
    \toprule[1.2pt]
    \multirow{2}{*}{\textbf{Method}} & \multirow{2}{*}{\bf Attribute} & \multirow{2}{*}{\bf Scene} & \multicolumn{3}{c}{\bf Relation} & \multirow{2}{*}{\bf Overall}   \\
    \cmidrule{4-6}
    &  &  &  Spatial & Action & Part   \\
    \midrule
    CLIPScore (ViT-L-14) & 13.0 & 40.0 & 8.5 & 11.1 & 11.5 & 9.9 \\
    \midrule
    VQAScore (InstructBLIP)  & 52.2 & 70.0 & 41.4 & {\bf 50.0} & 50.0 & 48.1 \\ %
    VQAScore (LLaVA-1.5)  & 53.6 &  {\bf 80.0} & 47.6 & 27.8 & 57.7 & 47.3  \\
    \midrule
    {\bf VQAScore (CLIP-FlanT5)}  & {\bf 59.4} & {\bf 80.0} & {\bf 57.3} & 44.4 & {\bf 69.2} & {\bf 57.2} \\
    \bottomrule[1.2pt]
    \end{NiceTabular}
    }
         & 
         
             \begin{NiceTabular}{lccccc|c}
        \CodeBefore
          \rectanglecolor{softgreen}{4-1}{7-7}
        \Body
    \toprule[1.1pt]
    \multirow{2}{*}{\textbf{Method}} &  \multirow{2}{*}{\bf Count} & \multirow{2}{*}{\bf Differ} & \multirow{2}{*}{\bf Compare} & \multicolumn{2}{c}{\bf Logical}  & \multirow{2}{*}{\bf Overall}   \\
    \cmidrule{5-6}
    &  &  & &  Negate & Universal   \\
    \midrule
    CLIPScore (ViT-L-14) & 7.8 & 2.3 & 2.0 & 0.0 & 0.0 & 4.4\\
    \midrule
    VQAScore (InstructBLIP) & 37.3 & 11.6 & 22.4 & 40.0 & 0.0 & 20.4  \\ %
    VQAScore (LLaVA-1.5) & 29.4 & 20.9 & 16.3 & 40.0 & 0.0 & 24.1  \\
    \midrule
    {\bf VQAScore (CLIP-FlanT5)} & {\bf 54.9} &  {\bf 44.2} & {\bf 49.0} & {\bf 60.0} & {\bf 73.3}  & {\bf 51.1} \\
    \bottomrule[1.1pt]
    \end{NiceTabular} 
    
    \\
  {\bf (a) Basic skills} (excluding samples requiring advanced skills)   & {\bf (b) Advanced skills} (including samples requiring basic skills)
\end{tabular}
}
\label{tab:winoground_analysis}
\end{table*}

{\bf Evaluating VQAScore's agreement with human judgments.} We now test VQAScore on five text-to-image evaluation benchmarks (TIFA160~\cite{tifa}, Pick-a-Pic~\cite{pickscore}, and DrawBench~\cite{imagen}, EditBench~\cite{wang2023imagen}, COCO-T2I~\cite{coco}) to measure its correlation (or agreement) with human judgments of alignment. In these benchmarks, given a text prompt, humans rate each generated image on a 1-to-5 Likert scale or assign a binary match-or-not label. Additionally, we report on an image-to-text evaluation benchmark Flickr8K~\cite{clipscore}, where each caption is manually rated based on the image. We follow SeeTrue~\cite{vq2} to report AUROC on DrawBench, EditBench, and COCO-T2I. For TIFA160 and Flickr8K, we evaluate pairwise accuracy as advocated by Deutsch et al.~\cite{deutsch2023ties} (EMNLP'23 outstanding paper), since the original Kendall metric cannot handle ties common in human ratings. We report other metrics (e.g., Pearson and Kendall) in \appref{sec:eval_details}. Due to the excessive noisy labels and NSFW content in the original Pick-a-pic dataset~\cite{pickscore}, we manually filter its testset, resulting in a clean subset of 100 samples (each has one prompt and two images) for evaluating binary accuracy. 

{\bf VQAScore shows superior correlation with human judgments.} \autoref{tab:vqascore_results_seetrue_tifa_pick} shows that VQAScore sets a new SOTA across all text-to-image alignment benchmarks, outperforming methods that finetune using costly human feedback~\cite{pickscore, imagereward, hpsv2} or rely on proprietary models~\cite{viescore, vq2}. \appref{sec:eval_details} also shows that VQAScore achieves a new SOTA on the image-to-text alignment benchmark Flickr8K, outperforming methods like CIDEr and RefCLIPScore that require additional reference captions~\cite{clipscore}. Lastly, we highlight that the text prompts in these benchmarks lack the advanced compositional structure of real-world prompts (e.g., Winoground~\cite{winoground}). This motivates us to develop a benchmark with more challenging and realistic text prompts, which we present in the next \autoref{sec:genai_bench}.

\begin{table*}[h!]
\centering
\renewcommand{\arraystretch}{1.3}
\caption{\textbf{VQAScore on image-text alignment benchmarks that score agreement with human judgments of alignment.} We show AUROC for DrawBench, EditBench, and COCO-T2I; pairwise accuracy~\cite{deutsch2023ties} for TIFA160; and binary accuracy for Pick-a-Pick, with higher scores indicating better performance for all metrics. VQAScore (with our CLIP-FlanT5) outperforms all prior art across all benchmarks. In general, we find texts in these alignment benchmarks to lack the compositional structure compared to user-written prompts in benchmarks like Winoground~\cite{winoground}, motivating us to create GenAI-Bench.} 
\scalebox{0.68}{
\begin{NiceTabular}{llccccc}
            \CodeBefore
              \rectanglecolor{softgray}{12-1}{16-7}
              \rectanglecolor{softgreen}{17-1}{21-7}
            \Body
\toprule[1.5pt]
\multirow{1}{*}{\textbf{Method}}  & \multirow{1}{*}{\textbf{Models}} &  \multirow{1}{*}{{\bf DrawBench}}  & \multirow{1}{*}{{\bf EditBench}} & \multirow{1}{*}{{\bf COCO-T2I}} &
\multirow{1}{*}{\textbf{TIFA160}} & \multirow{1}{*}{\textbf{Pick-a-Pic}}
\\ 
\midrule
\multicolumn{3}{l}{\textit{{Based on vision-language models}}} \\
\multirow{1}{*}{CLIPScore~\cite{clipscore}}  & CLIP-L-14 & 49.1 & 60.6 & 63.7  & 54.1 & 76.0 \\
\multirow{1}{*}{BLIPv2Score~\cite{blipv2}} & BLIPv2 & 60.5 & 68.0 & 70.7 & 57.5  & 80.0 \\ \midrule
\multicolumn{3}{l}{\textit{Finetuned on human feedback}} \\
\multirow{1}{*}{PickScore~\cite{pickscore}} & CLIP-H-14 (finetuned) & 72.3 & 64.3 & 61.5 & 59.4  & 70.0  \\
\multirow{1}{*}{ImageReward~\cite{imagereward}} & BLIPv2 (finetuned) & 70.4 & 70.3 & 77.0 & 67.3 & 75.0 \\
\multirow{1}{*}{HPSv2~\cite{hpsv2}} & CLIP-H-14 (finetuned) & 63.1 & 64.1 & 60.3 & 55.2 & 69.0 \\ \midrule
\multicolumn{4}{l}{\textit{Divide-and-conquer via VQA}} \\
\multirow{1}{*}{VQ2~\cite{vq2}} & FlanT5, LLaVA-1.5 & 52.8 & 52.8 & 47.7 & 48.7 & 73.0\\
\multirow{1}{*}{Davidsonian~\cite{davidsonian}} & ChatGPT, LLaVA-1.5 & 78.8 & 69.0 & 76.2 & 54.3 & 70.0 \\ \midrule
\multicolumn{3}{l}{\textit{Based on proprietary models}} \\ 
\multirow{1}{*}{TIFA~\cite{tifa, vq2}} & Llama-2, PaLI-17B & 73.4 & 67.8 & 72.0 & -- & -- \\
VQ2~\cite{vq2} & FlanT5, PaLI-17B & 82.6 & 73.6 & 83.4 & -- & -- \\
GPT4V-Eval~\cite{zhang2023gpt} & GPT4-Vision & -- & -- & -- & 64.0  & 74.0 \\
VIEScore~\cite{viescore} & GPT4-Vision & -- & -- & -- & 63.9 & 78.0 \\ \midrule
\multicolumn{4}{l}{\it VQAScore (ours) using open-source VQA models} \\
\multirow{1}{*}{\bf VQAScore} & InstructBLIP & 82.6 & 75.7 & 83.0 & 70.1 & 83.0 \\ %
\multirow{1}{*}{\bf VQAScore} & LLaVA-1.5 & 82.2 & 70.6 & 79.4 & 66.4 & 76.0 \\ %
\midrule
\multicolumn{4}{l}{\textit{VQAScore (ours) using our VQA model}} \\
\multirow{1}{*}{\bf VQAScore} & {\bf CLIP-FlanT5 (Ours)} & {\bf 85.3} &	{\bf 77.0} &	{\bf 85.0} &	{\bf 71.2}  &	{\bf 84.0} \\ %
\bottomrule[1.5pt]
\end{NiceTabular}
}
\label{tab:vqascore_results_seetrue_tifa_pick}
\end{table*}

\section{GenAI-Bench for Text-to-Visual Evaluation}
\label{sec:genai_bench}
In this section, we introduce {\bf GenAI-Bench}, a more challenging benchmark with compositional text prompts to evaluate both (1) text-to-visual generation models and (2) vision-language alignment metrics. Below, we present a preliminary study on GenAI-Bench, with further analysis in \cite{li2024evaluating, li2024genaibench}.%

{\bf Collecting GenAI-Bench.} Inspired by the compositional structure of real-world (user-written) prompts~\cite{winoground, midjourney}, GenAI-Bench gathers text prompts covering essential visio-linguistic compositional reasoning skills, especially advanced ones (e.g., comparison, counting, logic) that are not fully explored in previous benchmarks, e.g., PartiPrompt~\cite{parti}, DrawBench~\cite{imagen}, and T2I-CompBench~\cite{huang2023t2i}. First, we collaborate with graphic designers who routinely use text-to-image tools like Midjourney~\cite{midjourney} to compile a comprehensive set of skills by surveying recent benchmarks~\cite{parti, imagen, huang2023t2i} and real-world prompts~\cite{midjourney}. Next, we collect prompts from these designers and ensure the prompts are relevant for real-world usage and free from subjective or toxic content, e.g., malicious web users often craft prompts with NSFW content~\cite{pickscore}. \appref{sec:genai_bench_details} discusses the issues we found in previous benchmarks~\cite{parti, pickscore, huang2023t2i}. Lastly, we carefully tag each prompt with {\em all} its associated visio-linguistic skills, in contrast to previous benchmarks that either release no tags~\cite{hpsv2, pickscore, evalcrafter} or limit them to one or two~\cite{parti, imagen, huang2023t2i}. The final GenAI-Bench contains 1,600 text prompts with over 5,000 human-verified skill tags. \appref{sec:skill_details} details the skill definitions and \appref{sec:genai_bench_details} describes the collection procedure.

\begin{figure*}[h!]
\centering
    \scalebox{0.79}{
        \begin{tabular}{c@{\hspace{4pt}}@{\hspace{4pt}}c}
           (a) Examples of GenAI-Bench that challenge top generative models  &  (b) GenAI-Bench leaderboard  \\
           \includegraphics[height=0.25\textheight]{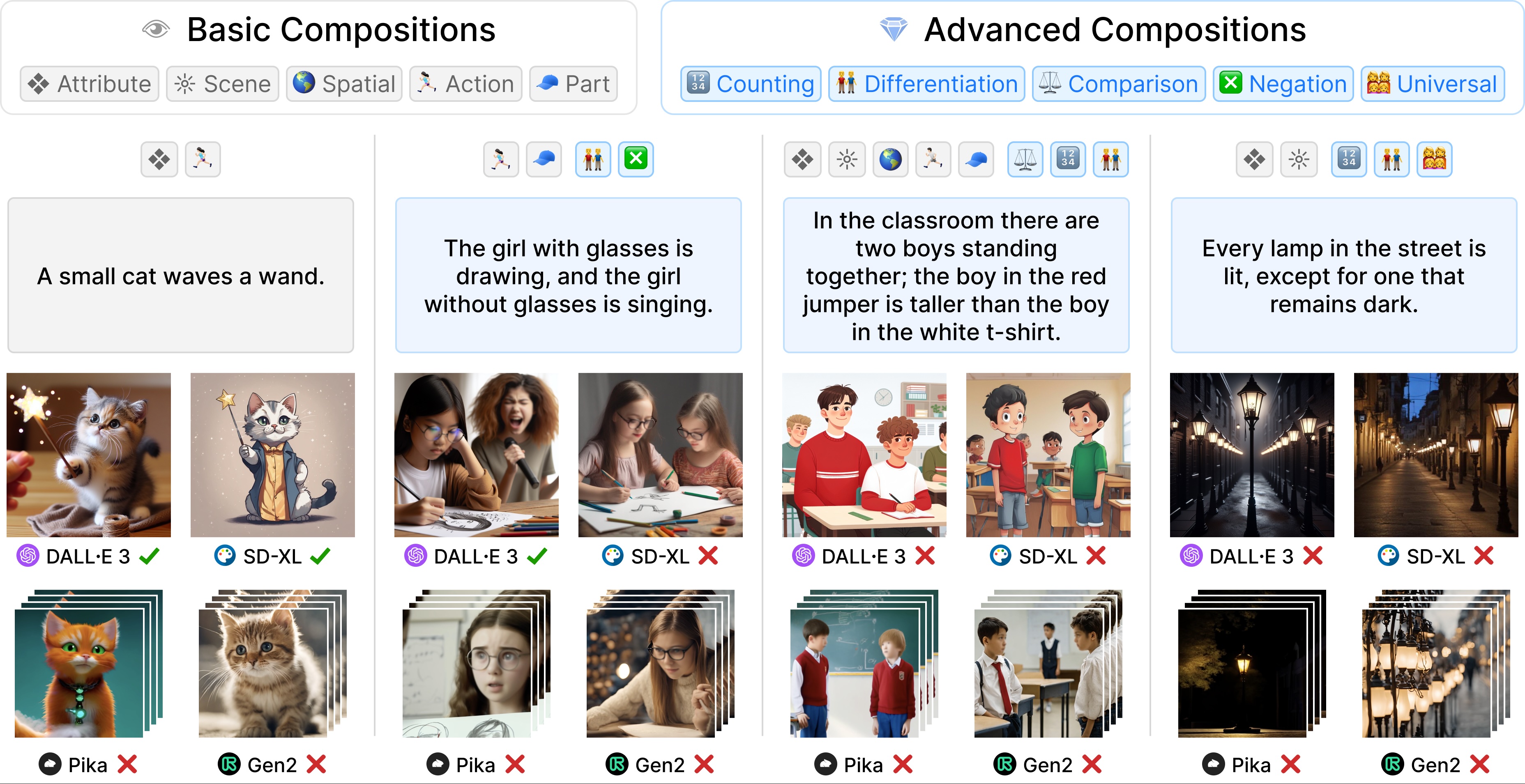} & \includegraphics[height=0.25\textheight]{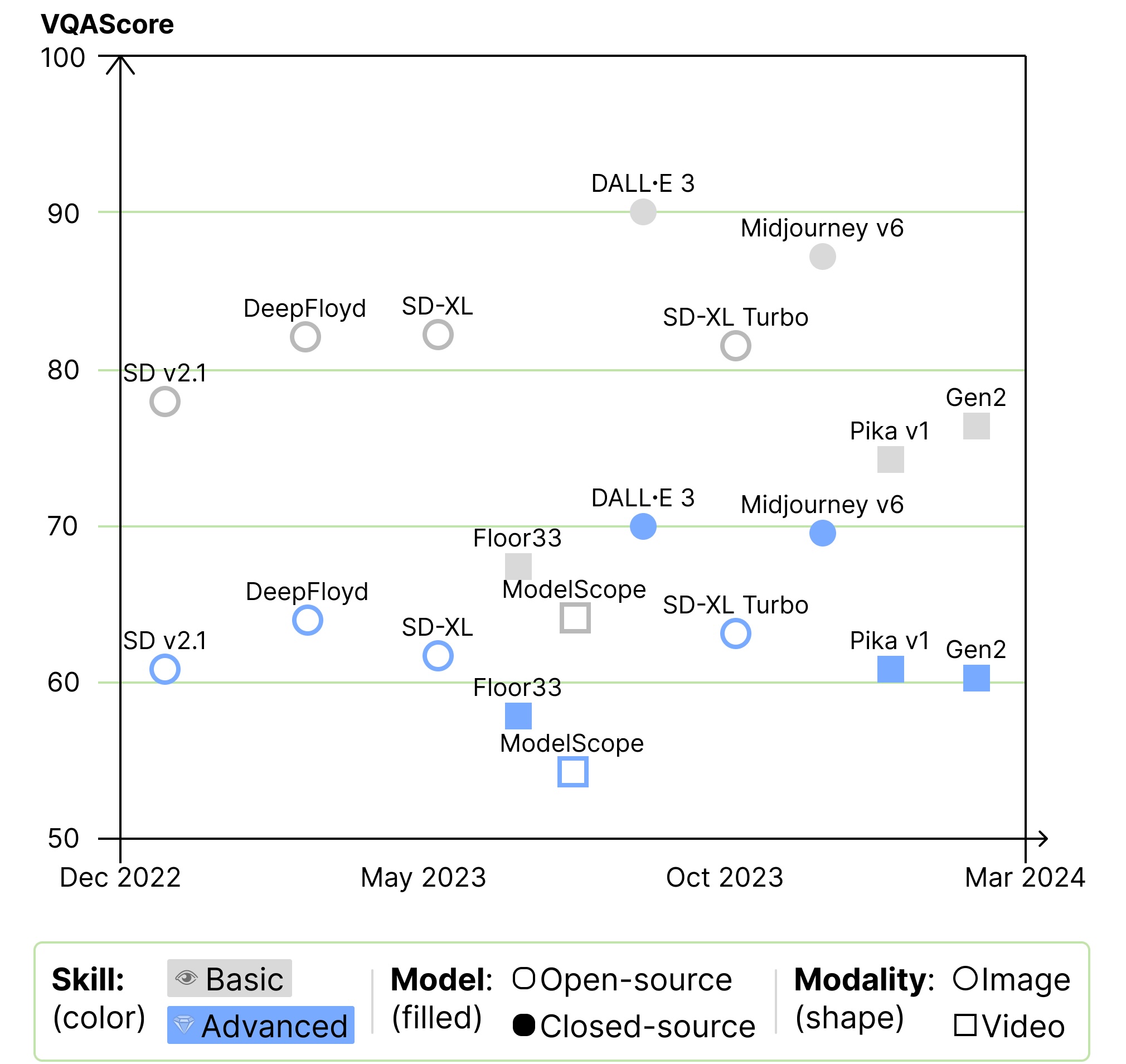}
        \end{tabular}
    }
    \caption{\small {\bf GenAI-Bench.} {\bf Figure (a)} shows example prompts and associated skill tags from GenAI-Bench. The advanced compositional prompts of GenAI-Bench pose greater challenges to leading image and video generative models. {\bf Figure (b)} presents the GenAI-Bench performance of 10 open/closed-source generative models. For each model, we separately show the averaged VQAScore for basic (in \textcolor{gray}{gray}) and advanced (in \textcolor{blue}{blue}) prompts. We find that (1) ``advanced'' prompts challenge all models more, (2) models that use stronger text embeddings or captions (e.g., DALL-E 3~\cite{dalle3} and DeepFloyd~\cite{deepfloyd}) achieve the best results, (3) open-source and video generative models~\cite{stablediffusion, gen2} still lag behind their closed-source and image counterparts~\cite{dalle3, midjourney}, indicating potential for further improvement. \autoref{sec:genai_bench_details} confirms that VQAScore agrees with collected human ratings.}
    \label{fig:genai_bench_leaderboard}
\end{figure*}

{\bf GenAI-Bench challenges leading text-to-visual models.} 
\autoref{fig:genai_bench_leaderboard}-a shows that state-of-the-art image and video generative models, such as DALL-E 3~\cite{dalle3}, Stable Diffusion (SD-XL)~\cite{stablediffusion}, Pika~\cite{pikalab}, and Gen2~\cite{gen2}, struggle with GenAI-Bench's compositional text prompts that require higher-order reasoning such as comparison, differentiation, counting, and logic. \autoref{fig:genai_bench_leaderboard}-b compares the averaged VQAScore (based on CLIP-FlanT5) of six image and four video generative models~\footnote{This work uses a coreset of 527 prompts for both human and automated evaluation.}. We compute VQAScore for video-text pairs by averaging across all video frames as described in \autoref{sec:video_and_3d}. We separately analyze each model's performance on ``basic'' and ``advanced'' prompts. Our analysis reveals significant improvements in text-to-visual generation for ``basic'' prompts from 2022 to 2023; however, improvements are less pronounced for ``advanced'' prompts, reflected in lower VQAScores across models. Nonetheless, we find that models with stronger language capabilities generally perform better. For example, one of the best open-source models DeepFloyd-IF~\cite{deepfloyd} uses strong text embeddings from the T5 language model~\cite{t5} rather than CLIP's, which do not encode compositional structure~\cite{kamath2023text}. Similarly, the best closed-source model DALL-E 3~\cite{dalle3} does not directly train on noisy web text captions but instead improves them using captioning models. Finally, we anticipate significant advancements in open-source and video-generative models (e.g., SD-XL~\cite{stablediffusion} and Gen2~\cite{gen2}), which currently lag behind their closed-source and image-generative counterparts.

{\bf VQAScore agrees with human judgments on GenAI-Bench.} We hire three annotators to rate the image-text (or video-text) pairs on a 1-5 Likert scale, following the annotation protocol of \cite{otani2023toward}. In this work, we report on a core set of 527 prompts and collect 15,810 ratings across the ten generative models, significantly exceeding the scale of human evaluation in previous work~\cite{tifa, imagen}. We extend our analysis to all 1,600 prompts in \cite{li2024evaluating}. \appref{sec:genai_bench_details} confirms that VQAScore achieves the state-of-the-art correlation with human judgments on GenAI-Bench. We will release these human ratings to support the development of future alignment metrics.

\section{Extending VQAScore to Video and 3D}
\label{sec:video_and_3d}
We now show that VQAScore can evaluate the alignment of text-to-video and 3D models.

{\bf Video and 3D alignment benchmarks.} For video-text alignment, we use the EvalCrafter~\cite{evalcrafter} benchmark with 1-5 human Likert scales collected by \cite{t2vscore}. For 3D-text alignment, we adopt the StanfordT23D benchmark~\cite{wu2024gpt}, which released 3D assets but not human ratings. As such, we collect over 1,000 human 1-5 Likert scales on six text-to-3D models. We report Pairwise accuracy~\cite{deutsch2023ties}, Pearson, and Kendall on both benchmarks. 

{\bf VQAScore achieves SOTA on video/3D-text alignment.} To compute VQAScore using VQA models trained solely on images, we uniformly sample video frames across time and 2D views from 3D assets across camera angles. \autoref{tab:text_to_video_3d}-a shows that our VQAScore  surpasses the divide-and-conquer approach T2VScore-A~\cite{t2vscore} based on GPT4-Vision. \autoref{tab:text_to_video_3d}-b shows that VQAScore exceeds popular text-to-3D metrics~\cite{wu2024gpt} such as CLIPScore~\cite{clipscore} and PickScore~\cite{pickscore}. In \appref{sec:eval_details}, we show it is possible to achieve near-optimal performance using as few as 4 video frames (or 9 views), in contrast to the 36 video frames (or 120 views) provided by the original benchmarks. 

\begin{table*}[h!]
\centering
\renewcommand{\arraystretch}{1.3}
\caption{{\bf Evaluating VQAScore on text-to-video/3D benchmarks.} We uniformly sample frames from videos and rendered views from 3D assets to calculate the average VQAScore (and other metrics). We report Pairwise accuracy, Pearson, and Kendall, with higher scores indicating better performance for all metrics. VQAScore surpasses popular video/3D metrics like CLIPScore~\cite{clipscore}, PickScore~\cite{pickscore}, and methods based on the proprietary GPT4-Vision~\cite{t2vscore} on both benchmarks.
}
\scalebox{0.96}{
\begin{tabular}{cc}
\scalebox{0.66}{
  \begin{NiceTabular}{lccc}
    \CodeBefore
      \rectanglecolor{softgray}{13-1}{14-4}
      \rectanglecolor{softgreen}{15-1}{19-4}
    \Body
\toprule[1.5pt]
\multirow{2}{*}{\textbf{Method}} & \multirow{2}{*}{\textbf{\small \begin{tabular}{@{}c@{}}Pairwise \\ Acc~\cite{deutsch2023ties}\end{tabular}}} & \multicolumn{2}{c}{\textbf{\small \textcolor{gray}{Old Metrics}}}  \\ \cmidrule(r){3-4} 
& & \textcolor{gray}{\small Pearson} & \textcolor{gray}{\small Kendall} \\
\midrule
\multicolumn{4}{l}{\textit{Baselines reported in \cite{t2vscore}}}\\
CLIPScore & 59.9 & \textcolor{gray}{34.3} & \textcolor{gray}{23.6} \\
X-CLIPScore & 56.9 & \textcolor{gray}{25.7} & \textcolor{gray}{17.5} \\
BLIP-BLEU & 53.0 & \textcolor{gray}{15.2} & \textcolor{gray}{10.4} \\
\midrule
\multicolumn{4}{l}{\textit{T2VScore-A reported in~\cite{t2vscore}}} \\
Otter-Video & -- & \textcolor{gray}{18.1} & \textcolor{gray}{13.4} \\
Video-LLaMA & -- & \textcolor{gray}{28.8} & \textcolor{gray}{20.6} \\
mPLUG-OWL2-Video  & -- & \textcolor{gray}{39.4} & \textcolor{gray}{28.5} \\
mPLUG-OWL2-Image & -- & \textcolor{gray}{35.8} & \textcolor{gray}{25.7} \\
InstructBLIP & -- & \textcolor{gray}{34.2} & \textcolor{gray}{24.6} \\
\midrule
\multicolumn{4}{l}{\textit{T2VScore-A w/ GPT4-V~\cite{t2vscore}}} \\
GPT4-Vision & 61.4 & \textcolor{gray}{48.6} & \textcolor{gray}{36.0} \\
\midrule
\multicolumn{4}{l}{\textit{VQAScore w/ open-source models}} \\
InstructBLIP & 65.8 & \textcolor{gray}{46.5} & \textcolor{gray}{35.8}  \\ %
LLaVA-1.5 & 63.7 & \textcolor{gray}{44.9} & \textcolor{gray}{31.4}   \\
\midrule
\multicolumn{4}{l}{\textit{VQAScore w/ our model}} \\
{\bf CLIP-FlanT5 (Ours)} & {\bf 66.5} & \textcolor{gray}{{\bf 49.1}} & \textcolor{gray}{{\bf 37.1}}   \\
\bottomrule[1.5pt]
\end{NiceTabular}   
}
& 
\scalebox{0.87}{
\begin{NiceTabular}{lccc}
    \CodeBefore
      \rectanglecolor{softgreen}{10-1}{14-4}
    \Body
\toprule[1.2pt]
\multirow{2}{*}{\textbf{Method}} & \multirow{2}{*}{\textbf{\small \begin{tabular}{@{}c@{}}Pairwise \\ Acc~\cite{deutsch2023ties}\end{tabular}}} & \multicolumn{2}{c}{\textbf{\small \textcolor{gray}{Old Metrics}}}  \\ \cmidrule(r){3-4} 
& & \textcolor{gray}{\small Pearson} & \textcolor{gray}{\small Kendall} \\
\midrule
\multicolumn{4}{l}{\textit{Baselines}} \\
CLIPScore~\cite{clipscore} & 61.0 & \textcolor{gray}{48.1} & \textcolor{gray}{32.6} \\
BLIPv2Score~\cite{blipv2} & 56.6 & \textcolor{gray}{34.3} & \textcolor{gray}{23.4}   \\
\midrule
\multicolumn{4}{l}{\textit{Finetuned on human feedback}} \\
ImageReward~\cite{imagereward} & 66.3 & \textcolor{gray}{57.1} & \textcolor{gray}{43.8}  \\ 
PickScore~\cite{pickscore} & 60.1  & \textcolor{gray}{41.3} & \textcolor{gray}{30.8}  \\ 
HPSv2~\cite{hpsv2} & 55.9 &  \textcolor{gray}{31.5} & \textcolor{gray}{21.9}   \\
\midrule
\multicolumn{5}{l}{\textit{VQAScore w/ open-source models}} \\
InstructBLIP & 68.0 & \textcolor{gray}{59.5} & \textcolor{gray}{47.5}  \\ %
LLaVA-1.5 & 64.9 & \textcolor{gray}{55.8} & \textcolor{gray}{40.8}   \\
\midrule
\multicolumn{5}{l}{\textit{VQAScore w/ our model}} \\
{\bf CLIP-FlanT5 (Ours)} & {\bf 68.6} & \textcolor{gray}{{\bf 64.3}} & \textcolor{gray}{{\bf 48.7}}   \\
\bottomrule[1.2pt]
\end{NiceTabular}
}

\\

\textbf{\scriptsize (a) Text-to-video benchmark (T2VScore~\cite{t2vscore})}  & \textbf{\scriptsize (b) Text-to-3D benchmark (StanfordT23D~\cite{wu2024gpt})} 
\end{tabular}
}
\vspace{-3mm}
\label{tab:text_to_video_3d}
\end{table*}

\section{Conclusion}
\label{sec:conclusion}
{\bf Limitations and future work.} While VQAScore excels in vision-language alignment, it currently does not evaluate other critical aspects of generative models~\cite{lee2023holistic, 
parashar2024neglected, evalcrafter, wu2024gpt}, such as toxicity, bias, aesthetics, video motion, and 3D physics. We posit that VQAScore can evaluate these aspects if it were finetuned on relevant data.

{\bf Summary.} We introduce VQAScore, a simple method surpassing current alignment metrics in evaluating text-to-image/video/3D models. VQAScore based on our CLIP-FlanT5 model offers a strong alternative to CLIPScore, especially on real-world compositional text prompts. We also introduce a more challenging GenAI-Bench to evaluate both text-to-visual generative models and vision-language alignment metrics. We hope our novel metric and benchmark will advance the scientific evaluation of generative models.

\newpage
\section{Acknowledgement}
\label{sec:acknowledgement}
We express our deepest gratitude to the Meta GenAI team (Xiaoliang Dai, Miao Liu, Peizhao Zhang, Peter Vajda, Ning Zhang) for supporting this work. We thank Harman Singh, Zihan Wang, Jean de Dieu Nyandwi, Simran Khanuja, Zixian Ma, and Ranjay Krishna for their invaluable discussions during the development of this work. We also thank Tiffany Ling for her contributions to the visual design.

{\small
\bibliographystyle{natbib}
\bibliography{main}
}

\newpage
\clearpage

{
    \centering
    \Large
    \textbf{Evaluating Text-to-Visual Generation \\ with Image-to-Text Generation} \\ \vspace{0.5em} {Supplementary Material}\\
    \large
}

\appendix

\vspace{-0.1in}

\section*{}
\begin{center}
    \emph{\bf \em \large Outline}
\end{center}
{This document supplements the main paper with benchmark and method details. Below is the outline:
\begin{itemize}
\item {\bf Section \ref{sec:skill_details}} details the skill definitions of GenAI-Bench and compares the skill coverage across popular benchmarks.
\item {\bf Section \ref{sec:genai_bench_details}} describes how GenAI-Bench is collected and shows VQAScore's strong agreement with human judgments.
\item {\bf Section \ref{sec:vqascore_details}} describes how we compute VQAScore with equations and pseudocode.
\item {\bf Section \ref{sec:clip_flant5_ablations}} includes the implementation details of CLIP-FlanT5 and ablation studies of training data, model size, and question-answer templates.
\item {\bf Section \ref{sec:method_details}} provides details on the baseline methods, including more failure cases of divide-and-conquer approaches.
\item {\bf Section \ref{sec:eval_details}} provides details on the benchmarks and evaluation metrics, and ablates sampling methods for video and 3D.

\end{itemize}
}

\section{Visio-Linguistic Compositional Reasoning Skills}
\label{sec:skill_details}
This section describes how we define and label the compositional reasoning skills for text-to-visual generation, and compare the skill coverage across benchmarks.

{\bf Skill definitions.} Prior literature on text-to-visual generation~\cite{parti, imagen, tifa, dalleval, huang2023t2i} focuses on generating ``basic'' objects, attributes, relations, and scenes. However, user prompts often require ``advanced'' compositional reasoning, including comparison, differentiation, counting, and logic~\cite{li2024evaluating, liu2022compositional}. For example, user prompts may require counting not just objects, but also attribute-object pairs and even object-relation-object triplets, like ``{\tt one person wearing a white shirt and the other five wearing blue shirts}''. To this end, after thoroughly reviewing relevant literature~\cite{winoground, midjourney, huang2023t2i, parti}, we work with professional designers to design a taxonomy of compositional reasoning skills common in real-world prompts, categorizing them into ``basic'' and ``advanced'', where the latter builds upon the former. We provide detailed definitions for ``basic'' skills in \autoref{tab:basic_skills} and ``advanced'' skills in \autoref{tab:advanced_skills}. 

\begin{table*}[h!]
\centering
\caption{\textbf{Skill definitions and examples for basic compositions.} }
\scalebox{0.8}{
\begin{NiceTabular}{l M{0.4\linewidth} M{0.5\linewidth}}
\CodeBefore
      \rectanglecolor{softgray}{2-1}{15-3}
    \Body
\toprule[1.2pt]
 \textbf{Skill Type} & \textbf{Definition} & \textbf{Examples} \\ \midrule
 \multicolumn{3}{c}{\bf Basic Compositions}\\
 {\fontfamily{cmtt}\selectfont Object} & {Basic entities within an image, such as person, animal, food, items, vehicles, or text symbols (e.g., ``A'', ``1+1'').} & {\it a {\bf dog}, a {\bf cat} and a {\bf chicken} on a {\bf table}; a young {\bf man} with a green {\bf bat} and a blue {\bf ball}; a '{\bf No Parking}' sign on a busy street.} \\\\
 {\fontfamily{cmtt}\selectfont Attribute} & {Visual properties of entities, such as color, material, emotion, size, shape, age, gender, state, and so on.} & {\it a {\bf silver} spoon lies to the left of a {\bf golden} fork on a {\bf wooden} table; a {\bf green} pumpkin is smiling {\bf happily}, a {\bf red} pumpkin is sitting {\bf sadly}.} \\\\
 {\fontfamily{cmtt}\selectfont Scene} & {Backgrounds or settings of an image, such as weather and location.} & {\it A child making a sandcastle on a {\bf beach in a cloudy day}; a grand fountain surrounded by historic buildings in a {\bf town square}.} \\\\
 {\fontfamily{cmtt}\selectfont Spatial Relation} & {Physical arrangements of multiple entities relative to each other, e.g., on the right, on top, facing, towards, inside, outside, near, far, and so on.} & {\it a bustling city street, a neon 'Open 24 Hours' sign glowing {\bf above} a small diner; a teacher standing {\bf in front of} a world map in a classroom; tea steams {\bf in} a cup, {\bf next to} a closed diary with a pen resting {\bf on} its cover.} \\\\
 {\fontfamily{cmtt}\selectfont Action Relation} & {Action interactions between entities, e.g., pushing, kissing, hugging, hitting, helping, and so on.} & {\it a dog {\bf chasing} a cat; a group of children {\bf playing} on the beach; a boat {\bf glides} across the ocean, dolphins {\bf leaping} beside it and seagulls {\bf soaring} overhead.} \\\\
 {\fontfamily{cmtt}\selectfont Part Relation} & {Part-whole relationships between entities -- one entity is a component of another, such as body part, clothing, and accessories.} & {\it a pilot {\bf with aviator sunglasses}; a baker {\bf with a cherry pin on a polka dot apron}.; a young lady {\bf wearing a T-shirt} puts {\bf her hand} on a {\bf puppy's head}.}  \\
\bottomrule[1.2pt]
\end{NiceTabular}
}
\label{tab:basic_skills}
\end{table*}

\begin{table*}[h!]
\centering
\caption{\textbf{Skill definitions and examples for advanced compositions.} }
\scalebox{0.8}{
\begin{NiceTabular}{l M{0.4\linewidth} M{0.5\linewidth}}
\CodeBefore
      \rectanglecolor{softblue}{2-1}{15-3}
    \Body
\toprule[1.2pt]
 \textbf{Skill Type} & \textbf{Definition} & \textbf{Examples} \\ \midrule
 \multicolumn{3}{c}{\bf Advanced Compositions}\\
 {\fontfamily{cmtt}\selectfont Counting} & {Determining the quantity, size, or volume of entities, e.g., objects, attribute-object pairs, and object-relation-object triplets.} & {\it {\bf two} cats playing with a {\bf single} ball; {\bf five} enthusiastic athletes and {\bf one} tired coach; {\bf one} pirate ship sailing through space, crewed by {\bf five} robots; {\bf three} pink peonies and {\bf four} white daisies in a garden.} \\\\
 {\fontfamily{cmtt}\selectfont Differentiation} & {Differentiating objects within a category by their attributes or relations, such as distinguishing between ``old'' and ``young'' people by age, or ``the cat on top of the table'' versus ``the cat under the table'' by their spatial relations.} & {\it {\bf one cat} is sleeping on the table and {\bf the other} is playing under the table; there are two men in the living room, {\bf the taller one} to the left of {\bf the shorter one}; a notebook lies open in the grass, with sketches on {\bf the left page} and blank space {\bf on the right}; there are two shoes on the grass, {\bf the one without laces} looks newer than {\bf the one with laces}.} \\\\
 {\fontfamily{cmtt}\selectfont Comparison} & {Comparing characteristics like number, attributes, area, or volume between entities.} & {\it there are {\bf more} people standing than sitting; between the two cups on the desk, the {\bf taller} one holds {\bf more} coffee than the {\bf shorter} one, which is half-empty; a small child on a skateboard has {\bf messier} hair than the person next to him; three little boys are sitting on the grass, and the boy in the middle looks the {\bf strongest}.} \\\\
 {\fontfamily{cmtt}\selectfont Negation} & {Specifying the absence or contradiction of elements, as indicated by ``no'', ``not'', or ``without'', e.g., entities not present or actions not taken.} & {\it  a bookshelf with {\bf no} books, only picture frames.; a person with short hair is crying while a person with long hair {\bf is not}; a smiling girl with short hair and {\bf no} glasses}; a cute dog {\bf without} a collar. \\\\
 {\fontfamily{cmtt}\selectfont Universality} & {Specifying when every member of a group shares a specific attribute or is involved in a common relation, indicated by words like ``every'', ``all'', ``each'', ``both''.} & {\it in a room, {\bf all} the chairs are occupied except one; a bustling kitchen where {\bf every} chef is preparing a dish; in a square, several children are playing, {\bf each} wearing a red T-shirt; a table laden with apples and bananas, where {\bf all} the fruits are green; the little girl in the garden has roses in {\bf both} hands.} \\
\bottomrule[1.2pt]
\end{NiceTabular}
}
\label{tab:advanced_skills}
\end{table*}

{\bf Comparing skills across benchmarks.} We find the skill categorization in benchmarks like PartiPrompt~\cite{parti} to be ambiguous or even confusing. For example, PartiPrompt introduces two categories ``{\em complex}'' and ``{\em fine-grained detail}''. The former refers to ``{\em ...fine-grained, interacting details or relationships between multiple participants}'', while the latter refers to ``{\em ...attributes or actions of entities or objects in a scene}''. Upon closer examination, the categorization of spatial, action, and part relations into these categories appears arbitrary. To address this, we compare the skill coverage across all alignment and generation benchmarks. For benchmarks (PartiPrompt/T2I-CompBench) with defined skill categories, we map their skills to our definitions. For benchmarks (Winoground/EqBen/Pick-a-pic/DrawBench/EditBench/COCO-T2I/HPDv2-Test/EvalCrafter) without a comprehensive skill set, we manually annotate the samples. Finally, we calculate the skill proportions in each benchmark, identifying skills that constitute more than 2\% as genuinely present. \autoref{tab:skill_comparison} shows that our GenAI-Bench comprehensively covers all essential skills in real-world prompts like those of \cite{winoground}.

\clearpage
\begin{table}[ht]
\centering
\caption{{\bf Comparing skill coverage across benchmarks.} Compared to existing alignment and generation benchmarks, GenAI-Bench comprehensively covers essential skills (especially advanced ones) in real-world prompts~\cite{midjourney} like those in Winoground~\cite{winoground}. Note that SeeTrue is an alignment benchmark proposed in \cite{vq2} that collects 6,930 human labels for DrawBench~\cite{imagen}, EditBench~\cite{wang2023imagen}, and COCO-T2I~\cite{coco}.}
\scalebox{0.7}{
    \begin{tabular}{@{}lcccccccccc@{}}
    \toprule
    \multirow{2}{*}{\bf Benchmarks} & \multicolumn{5}{c}{\bf Basic Compositions} & \multicolumn{5}{c}{\bf Advanced Compositions} \\ \cmidrule(l){2-6} \cmidrule(l){7-11}
     & Attribute & Scene & Action & Spatial & Part & Counting & Negation & Universal & Comparison & Differentiation\\ 
    \midrule
    \multicolumn{4}{l}{\textit{Alignment benchmarks}} &  &  &  &  &  &  &  \\ 
    Winoground~\cite{winoground} & \cmark & \cmark & \cmark & \cmark & \cmark & \cmark & \cmark & \cmark & \cmark  & \cmark \\  
    EqBen~\cite{eqben} & \cmark & \cmark & \cmark & \cmark & \cmark & \cmark & \cmark & \xmark & \xmark & \xmark  \\ 
    TIFA160~\cite{tifa} & \cmark & \cmark & \cmark & \cmark & \cmark & \cmark  & \xmark & \xmark & \xmark & \xmark  \\ 
    SeeTrue~\cite{vq2, wang2023imagen, imagen, coco} & \cmark & \cmark & \cmark & \cmark & \cmark & \cmark & \xmark & \xmark & \xmark & \xmark  \\ 
    Pick-a-pic~\cite{pickscore} & \cmark & \cmark & \cmark & \cmark & \cmark & \cmark & \xmark& \xmark & \xmark & \xmark  \\ 
    \midrule
    \multicolumn{4}{l}{\textit{Generation benchmarks}} &  &  &  &  &  &  &  \\ 
    PartiPrompt (P2)~\cite{parti} & \cmark & \cmark & \cmark & \cmark & \cmark & \cmark  & \cmark & \xmark & \xmark & \xmark  \\ 
    DrawBench~\cite{vq2, imagen} & \cmark & \cmark & \cmark & \cmark & \cmark & \cmark & \xmark & \xmark & \xmark & \xmark  \\ 
    EditBench~\cite{vq2, wang2023imagen} & \cmark & \cmark & \cmark & \cmark & \cmark & \cmark & \xmark & \xmark & \xmark & \xmark  \\ 
    COCO-T2I~\cite{vq2, coco} & \cmark & \cmark & \cmark & \cmark & \cmark & \cmark & \xmark & \xmark & \xmark & \xmark  \\ 
    T2I-CompBench~\cite{huang2023t2i} & \cmark & \cmark & \cmark & \cmark & \cmark & \cmark & \xmark & \xmark & \xmark & \xmark  \\ 
    HPDv2-Test~\cite{hpsv2} & \cmark & \cmark & \cmark & \cmark & \cmark & \xmark & \xmark & \xmark & \xmark & \xmark  \\ 
    EvalCrafter~\cite{evalcrafter} & \cmark & \cmark & \cmark & \cmark & \cmark & \cmark & \xmark & \xmark & \xmark & \xmark  \\ 
    \midrule \midrule
    \multicolumn{4}{l}{\textit{Our benchmark for both alignment and generation}} &  &  &  &  &  &  &  \\ 
    {\bf GenAI-Bench (Ours)}& \cmark & \cmark & \cmark & \cmark & \cmark & \cmark & \cmark & \cmark & \cmark  & \cmark \\ 
    \bottomrule
    \end{tabular}
}
\label{tab:skill_comparison}
\end{table}

\section{GenAI-Bench}
\label{sec:genai_bench_details}
This section describes how we collect GenAI-Bench and showcases VQAScore's superior agreement with human ratings.

{\bf Details of GenAI-Bench.} GenAI-Bench consists of 1,600 diverse prompts that cover advanced skills not addressed in previous benchmarks~\cite{parti,imagen,huang2023t2i}. To source prompts relevant to real-world applications, we employ two graphic designers who use Midjourney~\cite{midjourney} in their profession. First, we introduce them to our skill definitions and examples. Then, we ask them to craft prompts for each skill, collaborating with ChatGPT to brainstorm prompt variants across diverse visual domains. Importantly, these designers ensure that the prompts are {\em objective}. This contrasts with T2I-CompBench~\cite{huang2023t2i}, whose prompts are almost entirely auto-generated. For example, in T2I-CompBench's ``{\em texture}'' category, an overwhelming 40\% of the 1000 programmatically-generated prompts use ``metallic'' as the attribute, which limits their diversity. Other T2I-CompBench's prompts generated by ChatGPT often contain subjective (non-visual) phrases. For instance, in the prompt ``{\tt the delicate, fluttering wings of the butterfly signaled the arrival of spring, a natural symbol of rebirth and renewal}'', the ``rebirth and renewal'' can convey different meanings to different people. Similarly, in ``{\tt the soft, velvety texture of the rose petals felt luxurious against the fingertips, a romantic symbol of love and affection}'', the ``love and affection'' is also open to diverse interpretations. Thus, we carefully guide the designers to avoid such prompts. Lastly, each prompt in GenAI-Bench is tagged with its associated visio-linguistic skills. We streamline this process by using GPT4 for automatic tagging, providing it the skill definitions and in-context exemplars. Later, we manually verify and correct all tags for accuracy. This results in over 5,000 human-verified tags.

{\bf Collecting human ratings.} We evaluate six text-to-image models: Stable Diffusion~\cite{stablediffusion} (SD v2.1, SD-XL, SD-XL Turbo), DeepFloyd-IF~\cite{deepfloyd}, Midjourney v6~\cite{midjourney}, DALL-E 3~\cite{dalle3}; along with four text-to-video models: ModelScope~\cite{modelscope}, Floor33~\cite{floor33}, Pika v1~\cite{pikalab}, Gen2~\cite{gen2}. In this preliminary study, we use a coreset of 527 prompts from GenAI-Bench. This already exceeds the scale of human annotations in previous work~\cite{tifa, vq2}. We will extend our benchmark to all 1,600 prompts in a subsequent study. Due to the lack of APIs for Floor33~\cite{floor33}, Pika v1~\cite{pikalab}, and Gen2~\cite{gen2}, we manually download videos from their websites. We plan to release our codebase for automatically generating visuals with the rest of the models. Finally, we collect 1-5 Likert scale human ratings using the recommended annotation protocol of \cite{otani2023toward}:
\clearpage
\begin{mdframed}[linewidth=1pt, linecolor=black, leftmargin=3cm, rightmargin=3cm, backgroundcolor=gray!20, roundcorner=5pt]
\scriptsize
\textcolor{blue}{How well does the image (or video) match the description?} \\
\quad 1. Does not match at all.\\
\quad 2. Has significant discrepancies.\\
\quad 3. Has several minor discrepancies.\\
\quad 4. Has a few minor discrepancies.\\
\quad 5. Matches exactly.
\end{mdframed}

\noindent Our collected human ratings indicate a high level of inter-rater agreement, with Krippendorff's Alpha reaching 0.72 for image ratings and 0.70 for video ratings, suggesting substantial agreement~\cite{tifa}. Further, we show that VQAScore achieves the state-of-the-art correlation to human ratings in \autoref{tab:genai_bench_human_correlation}.

\begin{table*}[h!]
\centering
\renewcommand{\arraystretch}{1.3}
\caption{{\bf Evaluating VQAScore on GenAI-Bench.} We report Pairwise accuracy, Pearson, and Kendall, with higher scores indicating better performance for all metrics. VQAScore sets a new SOTA on both the image and video alignment benchmarks of GenAI-Bench (with 527 prompts each), significantly surpassing popular metrics like CLIPScore~\cite{clipscore} and PickScore~\cite{pickscore}.
}
\scalebox{0.9}{
\begin{tabular}{cc}
\scalebox{0.8}{
\begin{NiceTabular}{lccc}
    \CodeBefore
      \rectanglecolor{softgreen}{10-1}{14-4}
    \Body
\toprule[1.2pt]
\multirow{2}{*}{\textbf{Method}} & \multirow{2}{*}{\textbf{\small \begin{tabular}{@{}c@{}}Pairwise \\ Acc~\cite{deutsch2023ties}\end{tabular}}} & \multicolumn{2}{c}{\textbf{\small \textcolor{gray}{Old Metrics}}}  \\ \cmidrule(r){3-4} 
& & \textcolor{gray}{\small Pearson} & \textcolor{gray}{\small Kendall} \\
\midrule
\multicolumn{4}{l}{\textit{Baselines}} \\
CLIPScore~\cite{clipscore} & 52.2 & \textcolor{gray}{19.9} & \textcolor{gray}{14.5} \\
BLIPv2Score~\cite{blipv2} & 55.1 & \textcolor{gray}{25.0} & \textcolor{gray}{20.7}   \\
\midrule
\multicolumn{4}{l}{\textit{Finetuned on human feedback}} \\
ImageReward~\cite{imagereward} & 58.7 & \textcolor{gray}{39.2} & \textcolor{gray}{28.3}  \\ 
PickScore~\cite{pickscore} & 57.7  & \textcolor{gray}{36.3} & \textcolor{gray}{26.2}  \\ 
HPSv2~\cite{hpsv2} & 49.8 &  \textcolor{gray}{14.5} & \textcolor{gray}{10.0}   \\
\midrule
\multicolumn{5}{l}{\textit{VQAScore w/ open-source models}} \\
InstructBLIP & 62.4 & \textcolor{gray}{43.9} & \textcolor{gray}{36.0}  \\ %
LLaVA-1.5 & 62.1 & \textcolor{gray}{\bf 48.3} & \textcolor{gray}{35.6}   \\
\midrule
\multicolumn{5}{l}{\textit{VQAScore w/ our model}} \\
{\bf CLIP-FlanT5 (Ours)} & {\bf 63.3} & \textcolor{gray}{46.9} & \textcolor{gray}{{\bf 38.0}}   \\
\bottomrule[1.2pt]
\end{NiceTabular}
}

& 
\scalebox{0.8}{
\begin{NiceTabular}{lccc}
    \CodeBefore
      \rectanglecolor{softgreen}{10-1}{14-4}
    \Body
\toprule[1.2pt]
\multirow{2}{*}{\textbf{Method}} & \multirow{2}{*}{\textbf{\small \begin{tabular}{@{}c@{}}Pairwise \\ Acc~\cite{deutsch2023ties}\end{tabular}}} & \multicolumn{2}{c}{\textbf{\small \textcolor{gray}{Old Metrics}}}  \\ \cmidrule(r){3-4} 
& & \textcolor{gray}{\small Pearson} & \textcolor{gray}{\small Kendall} \\
\midrule
\multicolumn{4}{l}{\textit{Baselines}} \\
CLIPScore~\cite{clipscore} & 54.5 & \textcolor{gray}{26.4} & \textcolor{gray}{19.1} \\
BLIPv2Score~\cite{blipv2} & 55.6 & \textcolor{gray}{27.4} & \textcolor{gray}{21.5}   \\
\midrule
\multicolumn{4}{l}{\textit{Finetuned on human feedback}} \\
ImageReward~\cite{imagereward} & 61.0 & \textcolor{gray}{44.7} & \textcolor{gray}{32.7}  \\ 
PickScore~\cite{pickscore} & 56.8  & \textcolor{gray}{33.5} & \textcolor{gray}{24.0}  \\ 
HPSv2~\cite{hpsv2} & 51.6 &  \textcolor{gray}{18.5} & \textcolor{gray}{13.2}   \\
\midrule
\multicolumn{5}{l}{\textit{VQAScore w/ open-source models}} \\
InstructBLIP & 62.6 & \textcolor{gray}{46.9} & \textcolor{gray}{36.2}  \\ %
LLaVA-1.5 & 64.3 & \textcolor{gray}{\bf 54.0} & \textcolor{gray}{39.7}   \\
\midrule
\multicolumn{5}{l}{\textit{VQAScore w/ our model}} \\
{\bf CLIP-FlanT5 (Ours)} & {\bf 64.4} & \textcolor{gray}{{53.3}} & \textcolor{gray}{{\bf 39.9}}   \\
\bottomrule[1.2pt]
\end{NiceTabular}
}

\\

\textbf{\scriptsize (a) GenAI-Bench-527 
 (Image)}  & \textbf{\scriptsize (b) GenAI-Bench-527 (Video) } 
\end{tabular}
}
\label{tab:genai_bench_human_correlation}
\end{table*}

{\bf GenAI-Bench performance.} We analyze the performance of the ten generative models across all skills in \autoref{tab:performance_breakdown}. Both human ratings and VQAScores prefer DALL-E 3~\cite{dalle3} over the other models in nearly all skills except for negation. In addition, prompts requiring ``advanced'' compositions are rated significantly lower by both humans and VQAScores. Lastly, current video models do not perform as well as image models, suggesting room for improvement.

\begin{table*}[h!]
\centering
\renewcommand{\arraystretch}{1.3}
\caption{{\bf Performance breakdown on GenAI-Bench.} We present the averaged human ratings and VQAScores (based on CLIP-FlanT5) for ``basic'' and ``advanced'' prompts. Human ratings use a 1-5 Likert scale, and VQAScore ranges from 0 to 1, with higher scores indicating better performance for both. Generally, both human ratings and VQAScores favor DALL-E 3 over other models, with DALL-E 3 preferred across almost all skills except for negation. In addition, we find that video models receive significantly lower scores than image models. Overall, VQAScore closely matches human ratings.  %
}
\scalebox{0.57}{
\begin{tabular}{c@{\hspace{8pt}}c}

  \begin{NiceTabular}{lccccc|c}
        \CodeBefore
        \Body
    \toprule[1.2pt]
    \multirow{2}{*}{\textbf{Method}} & \multirow{2}{*}{\bf Attribute} & \multirow{2}{*}{\bf Scene} & \multicolumn{3}{c}{\bf Relation} & \multirow{2}{*}{\bf Overall}   \\
    \cmidrule{4-6}
    &  &  &  Spatial & Action & Part   \\
    \midrule
    \multicolumn{5}{l}{\textit{Image models}}\\
    SD v2.1 & 3.1 & 3.2 & 2.9 & 3.2 & 3.1 & 3.1\\
    SD-XL & 3.7 & 3.7 & 3.4 & 3.7 & 3.6 & 3.6\\
    SD-XL Turbo & 3.6 & 3.7 & 3.3 & 3.5 & 3.5 & 3.5 \\
    DeepFloyd-IF & 3.6 & 3.7 & 3.4 & 3.7 & 3.6 & 3.6\\
    Midjourney v6 & 3.9 & 3.9 & 3.7 & 4.0 & 4.0 & 3.9\\
    DALL-E 3 & 4.3 & 4.5 & 4.3 & 4.3 & 4.3 & 4.3\\
    \midrule
    \multicolumn{5}{l}{\textit{Video models}}\\
    ModelScope & 3.0 & 3.1 & 2.8 & 3.1 & 3.2 & 2.9\\
    Floor33 & 3.1 & 3.2 & 2.9 & 3.3 & 3.2 & 3.1\\
    Pika v1 & 3.3 & 3.5 & 3.1 & 3.3 & 3.3 & 3.2\\
    Gen2 & 3.4 & 3.6 & 3.3 & 3.6 & 3.5 & 3.5\\
    \bottomrule[1.2pt]
    \end{NiceTabular}
    
         & 
           \begin{NiceTabular}{lccccc|c}
        \CodeBefore
        \Body
    \toprule[1.2pt]
    \multirow{2}{*}{\textbf{Method}} & \multirow{2}{*}{\bf Attribute} & \multirow{2}{*}{\bf Scene} & \multicolumn{3}{c}{\bf Relation} & \multirow{2}{*}{\bf Overall}   \\
    \cmidrule{4-6}
    &  &  &  Spatial & Action & Part   \\
    \midrule
    \multicolumn{5}{l}{\textit{Image models}}\\
    SD v2.1 & 0.80 & 0.79 & 0.76 & 0.77 & 0.80 & 0.78\\
    SD-XL & 0.84 & 0.84 & 0.82 & 0.83 & 0.89 & 0.83\\
    SD-XL Turbo & 0.83 & 0.83 & 0.80 & 0.81 & 0.84 & 0.82\\
    DeepFloyd-IF & 0.83 & 0.85 & 0.80 & 0.82 & 0.89 & 0.83\\
    Midjourney v6 & 0.88 & 0.87 & 0.87 & 0.87 & 0.91 & 0.87 \\
    DALL-E 3 & 0.91 & 0.90 & 0.92 & 0.89 & 0.91 & 0.90 \\
    \midrule
    \multicolumn{5}{l}{\textit{Video models}}\\
    ModelScope & 0.67 & 0.68 & 0.65 & 0.64 & 0.71 & 0.65\\
    Floor33 & 0.69 & 0.70 & 0.65 & 0.66 & 0.69 & 0.67\\
    Pika v1 & 0.77 & 0.79 & 0.74 & 0.71 & 0.76 & 0.74\\
    Gen2 & 0.77 & 0.79 & 0.73 & 0.76 & 0.84 & 0.76\\
    \bottomrule[1.2pt]
    \end{NiceTabular}

    \\
  {\bf (a) Human ratings on ``basic'' prompts}   & {\bf (b) VQAScores on ``basic'' prompts} \\ \\

        \begin{NiceTabular}{lccccc|c}
        \CodeBefore
        \Body
    \toprule[1.1pt]
    \multirow{2}{*}{\textbf{Method}} &  \multirow{2}{*}{\bf Count} & \multirow{2}{*}{\bf Differ} & \multirow{2}{*}{\bf Compare} & \multicolumn{2}{c}{\bf Logical}  & \multirow{2}{*}{\bf Overall}   \\
    \cmidrule{5-6}
    &  &  & &  Negate & Universal   \\
    \midrule
    \multicolumn{5}{l}{\textit{Image models}}\\
    SD v2.1 & 2.4 & 2.5 & 2.3 & 2.9 & 3.0 & 2.7\\
    SD-XL & 2.5 & 2.6 & 2.5 & 2.7 & 3.5 & 2.8\\
    SD-XL Turbo & 2.5 & 2.8 & 2.4 & 3.0 & 3.4 & 2.8\\
    DeepFloyd-IF & 2.8 & 2.9 & 2.6 & 2.9 & 3.6 & 3.0\\
    Midjourney v6 & 3.2 & 3.3 & 3.2 & 2.9 & 3.9 & 3.2\\
    DALL-E 3 & 3.3 & 3.4 & 3.4 & 2.8 & 4.0 & 3.3\\
    \midrule
    \multicolumn{5}{l}{\textit{Video models}}\\
    ModelScope & 2.1 & 2.3 & 2.0 & 2.7 & 3.0 & 2.5\\
    Floor33 & 2.6 & 2.8 & 2.4 & 3.0 & 3.4 & 2.8\\
    Pika v1 & 2.5 & 2.7 & 2.4 & 3.0 & 3.6 & 2.9\\
    Gen2 & 2.5 & 2.8 & 2.4 & 3.1 & 3.5 & 2.9\\
    \bottomrule[1.1pt]
    \end{NiceTabular}

         & 
         
    \begin{NiceTabular}{lccccc|c}
        \CodeBefore
        \Body
    \toprule[1.1pt]
    \multirow{2}{*}{\textbf{Method}} &  \multirow{2}{*}{\bf Count} & \multirow{2}{*}{\bf Differ} & \multirow{2}{*}{\bf Compare} & \multicolumn{2}{c}{\bf Logical}  & \multirow{2}{*}{\bf Overall}   \\
    \cmidrule{5-6}
    &  &  & &  Negate & Universal   \\
    \midrule
    \multicolumn{5}{l}{\textit{Image models}}\\
    SD v2.1 & 0.68 & 0.70 & 0.68 & 0.54 & 0.64 & 0.62\\
    SD-XL & 0.71 & 0.73 & 0.69 & 0.50 & 0.66 & 0.63\\
    SD-XL Turbo & 0.72 & 0.74 & 0.70 & 0.52 & 0.65 & 0.65\\
    DeepFloyd-IF & 0.74 & 0.74 & 0.71 & 0.53 & 0.68 & 0.66\\
    Midjourney v6 & 0.78 & 0.78 & 0.79 & 0.50 & 0.76 & 0.69\\
    DALL-E 3 & 0.82 & 0.78 & 0.82 & 0.48 & 0.80 & 0.70\\
    \midrule
    \multicolumn{5}{l}{\textit{Video models}}\\
    ModelScope & 0.56 & 0.61 & 0.56 & 0.51 & 0.55 & 0.55\\
    Floor33 & 0.66 & 0.69 & 0.61 & 0.53 & 0.56 & 0.58\\
    Pika v1 & 0.65 & 0.67 & 0.63 & 0.56 & 0.68 & 0.62\\
    Gen2 & 0.71 & 0.69 & 0.65 & 0.53 & 0.61 & 0.61\\
    \bottomrule[1.1pt]
    \end{NiceTabular} 
    
    \\
  {\bf (c) Human ratings on ``advanced'' prompts}   & {\bf (d) VQAScores on ``advanced'' prompts} 
\end{tabular}
}
\label{tab:performance_breakdown}
\end{table*}

\clearpage
\section{Implementing VQAScore}
\label{sec:vqascore_details}
In this section, we describe how we compute VQAScore.

{\bf Computing VQAScore as an auto-regressive product.} Recall that VQAScore calculates the alignment score of an image \textbf{i} and text \textbf{t} directly from a VQA model. We first use a simple QA template to convert the text {\bf t} to a question and an answer (denoted as {\bf q}({\bf t}) and {\bf a}({\bf t})), for example: 
\begin{align*}
\mathbf{t} &= \parbox[t]{.65\textwidth}{The moon is over the cow} \\
\mathbf{q}(\mathbf{t}) &= \parbox[t]{.65\textwidth}{Does this figure show "The moon is over the cow"? Please answer yes or no.} \\
\mathbf{a}(\mathbf{t}) &= \text{Yes}
\end{align*}

\noindent We later demonstrate that such a straightforward question-answer pair is sufficient for good performance. In language modeling~\cite{bengio2003neural}, a piece of text is pre-processed (or tokenized) into a token sequence, e.g., $\mathbf{a}(\mathbf{t}) = \{a_1, \cdots, a_m\}$. Although ``{\tt Yes}'' usually counts as a single token, we include the EOS (end-of-sentence) token at the end of the text sequence for a simpler implementation. We find that the EOS token only marginally affects the VQAScore results. Next, the generative likelihood of the answer (conditioned on both the question and image) can be naturally factorized as an auto-regressive product~\cite{bengio2003neural}:  

\begin{align}
\text{VQAScore}(\mathbf{i}, \mathbf{t}) := P(\mathbf{a}(\mathbf{t})|\mathbf{i},\mathbf{q}(\mathbf{t})) = \prod_{k=1}^m P(a_k | a_{<k}, \mathbf{i}, \mathbf{q}(\mathbf{t}))
\label{eq:product_of_conditionals}
\end{align}

\noindent The answer decoders of VQA models~\cite{llava, instructblip} return back $m$ softmax distributions corresponding to the $m$ terms in the above expression. Computing VQAScore is more efficient than generating answer token-by-token. Since the entire sequence of tokens $\{a_k\}$ is already available as input for VQAScore, the above $m$ terms can be efficiently computed in {\em parallel}. In contrast, answer generation as done by \cite{tifa, davidsonian} requires {\em sequential} token-by-token prediction, as token $a_k$ must be generated before it can serve as input to generate the softmax distribution for the subsequent token $a_{k+1}$. 

\begin{algorithm}[t]
\scriptsize
\SetAlgoLined
    \PyComment{tokenize(): text tokenizer that converts texts to a list of token indices} \\
    \PyComment{vqa\_model(): VQA model returns logits for predicted answer} \\
    \PyCode{} \\
    \PyCode{def vqa\_score(image, text):} \\
    \Indp
    
    \PyComment{Format the text into the below QA pair} \\
    \PyCode{question = f"Does this figure show `\{text\}'? Please answer yes or no."}  \\
    \PyCode{answer = "Yes"} \\
    \PyCode{} \\
    \PyComment{Tokenize the QA pair into tokens} \\
    \PyCode{question\_tokens = tokenize(question)} \\
    \PyCode{answer\_tokens = tokenize(answer)} \\
    \PyCode{} \\
    \PyComment{Extract logits for predicted answer of shape [len(answer\_tokens), vocab\_size]} \\
    \PyComment{answer\_tokens is a required input for auto-regressive decoding} \\
    \PyCode{logits = vqa\_model(image, question\_tokens, answer\_tokens)}\\ 

    \PyCode{} \\
    \PyComment{labels must skip the first BOS (Begin-Of-Sentence) token} \\
    \PyCode{labels = answer\_tokens[1:]} \\ 
    \PyComment{logits must skip the last EOS (End-Of-Sentence) token} \\
    \PyCode{logits = logits[:-1]}\\ 
    \PyCode{} \\
    \PyComment{Compute the log likelihood of the answer} \\
    \PyCode{log\_likelihood = -torch.nn.CrossEntropyLoss()(logits, labels)}\\
    \PyComment{(Optional) Cancel the log to obtain P("Yes" | image, question)} \\
    \PyCode{score = log\_likelihood.exp()}\\ 
    \PyCode{return score} \\
    
    \Indm
\caption{\footnotesize PyTorch-style pseudocode for VQAScore. 
}
\label{algo:vqascore}
\end{algorithm}

{\bf Pseudocode of VQAScore.} To better explain how VQAScore works, we attach the pseudocode in \autoref{algo:vqascore}. We will release a pip-installable API to compute VQAScore using one-line of Python code.

\section{Training CLIP-FlanT5}
\label{sec:clip_flant5_ablations}
In this section, we detail the training procedure of CLIP-FlanT5, and ablate design choices including training data, model size, and prompting strategies.

{\bf Training CLIP-FlanT5.} For a fair comparison, we adhere to the training recipe of the state-of-the-art LLaVA-1.5~\cite{llava15}. We adopt the same (frozen) CLIP visual encoder (ViT-L-336)~\cite{clip} and the 2-layer MLP projector for image tokenization. We also follow LLaVA-1.5's two-stage finetuning procedure and datasets. In stage-1 training, we finetune the MLP projector on 558K captioning data (LAION-CC-SBU
with BLIP captions~\cite{blipv2}). To accommodate FlanT5's encoder-decoder architecture, we adopt the split-text training method proposed in BLIPv2~\cite{blipv2}. This involves splitting a caption into two parts at a random position, with the first part sent to the encoder and the second part to the decoder. In stage-2 training, we finetune both the MLP projector and the language model (FlanT5) on 665K mixture of public VQA datasets (e.g., VQAv2~\cite{vqa2} and GQA~\cite{gqa}). To efficiently train the encoder-decoder architecture, we convert all multi-turn VQA samples into single-turn, resulting in 3.4M image-question-answer pairs. We also retrain LLaVA-1.5 on the same single-turn VQA samples and observe the same VQAScore results. We borrow hyperparameters of LLaVA-1.5 (see \autoref{tab:hyper}), such as the learning rate schedule, optimizer, number of epochs, and weight decay. We use 8 A100 (80Gbs) GPUs to train all our models. Our largest CLIP-FlanT5-XXL (11B) takes 5 hours for the stage-1 and 80 hours for the stage-2. For stage-2 training, we adhere to the system (prefix) prompt of LLaVA-1.5 during training~\footnote{By default, we also use the system prompt during inference. Interestingly, removing the system prompt (``A chat between a curious user ... answers to the user's questions'') during inference does not affect CLIP-FlanT5 but will hurt LLaVA-1.5's performance.}:

\begin{mdframed}[linewidth=1pt, linecolor=black, leftmargin=0.5cm, rightmargin=0.5cm, backgroundcolor=gray!20, roundcorner=5pt]
\scriptsize
A chat between a curious user and an artificial intelligence assistant. The assistant gives helpful, detailed, and polite answers to the user's questions. \\
\textbf{USER:} \textcolor{blue}{<image>} $\backslash$n \textcolor{red}{<question>} \textbf{ASSISTANT:} \textcolor{violet}{<answer>}
\end{mdframed}

\begin{table}[h]
\centering
\caption{\bf Training hyperparameters for CLIP-FlanT5.}
\scalebox{0.7}{
\begin{tabular}{l|cc}
\toprule[1pt]
\textbf{Hyperparameter} & \textbf{Stage-1} & \textbf{Stage-2} \\
\midrule
dataset size             & 558K             & 665K               \\
batch size             & 256               & 96               \\
lr                     & 1e-2              & 2e-5              \\
lr schedule            & \multicolumn{2}{c}{cosine decay}                       \\
lr warmup ratio        & \multicolumn{2}{c}{0.03}                                \\
weight decay           & \multicolumn{2}{c}{0}                                  \\
epoch                  & \multicolumn{2}{c}{1}                            \\
optimizer              & \multicolumn{2}{c}{AdamW} \\
DeepSpeed stage        & 2                 & 3                 \\
\bottomrule[1pt]
\end{tabular}
}
\label{tab:hyper}
\end{table}

{\bf Ablating language models and training data.} We evaluate four language models: the encoder-decoder FlanT5 (11B and 3B) and the decoder-only Llama-2 (13B and 7B). We also ablate finetuning strategies: using both captioning and VQA data (stage-2) against only captioning data (stage-1). We report overall performance across 7 image-text alignment benchmarks in \autoref{tab:training_ablation}. We highlight three key observations: 
\begin{enumerate}
    \item {\bf Finetuning on VQA data} is crucial (whereas captioning data only helps a little).
    \item {\bf Scaling up language models} consistently boosts performance.
    \item {\bf Encoder-decoder FlanT5} significantly outperforms decoder-only Llama-2.
\end{enumerate}

\noindent We hope our ablations can help future work develop stronger models for VQAScore. We will make all model checkpoints and data available for reproducibility. 

\begin{table*}[h!]
\centering
\renewcommand{\arraystretch}{1.3}
\caption{\textbf{Ablation on language model and training data.} We show overall performance on seven benchmarks: group score on Winoground/EqBen, AUROC on DrawBench/EditBench/COCO-T2I, pairwise accuracy on TIFA160, and binary accuracy on Pick-a-pic, with higher scores indicating better performance for all metrics. We highlight that scaling up the size of LLMs and finetuning on VQA data consistently improve the performance. In addition, the encoder-decoder FlanT5 is stronger than the decoder-only Llama-2, likely because FlanT5 benefits from bidirectional image-question encoding~\cite{ul2} and extensive training on challenging QA datasets~\cite{flant5}. } 
\scalebox{0.66}{
\begin{NiceTabular}{lllcccccccc}
            \CodeBefore
            \Body
\toprule[1.5pt]
\multirow{1}{*}{\textbf{LLM-Type}}  & \multirow{1}{*}{\textbf{Model-Size}} & \multirow{1}{*}{\textbf{Training-Data}} & \multirow{1}{*}{\textbf{Winoground}} & \multirow{1}{*}{\textbf{EqBen}} & \multirow{1}{*}{{\bf DrawBench}}  & \multirow{1}{*}{{\bf EditBench}} & \multirow{1}{*}{{\bf COCO-T2I}} &
\multirow{1}{*}{\textbf{TIFA160}} & \multirow{1}{*}{\textbf{Pick-a-Pic}}
\\ 
\midrule
\multirow{4}{*}{Llama-2} & \multirow{2}{*}{7B} & Caption Only & 3.8 & 7.9 & 42.5 & 45.0 & 46.2 & 46.6 & 53.0 \\ 
 &  & Caption+VQA & 21.8 & 20.7 & 81.7 & 65.6 & 80.5 & 64.9 & 81.0 \\ 
 \cmidrule{2-10}
 & \multirow{2}{*}{13B} & Caption Only & 0.8 & 1.4 & 56.5 & 47.0 & 51.5 & 49.7 & 44.0 \\ 
 & & Caption+VQA & 29.8 & 35.0 & 82.2 & 70.6 & 79.4 & 66.4 & 76.0 \\ 
\midrule
\multirow{4}{*}{FlanT5} & \multirow{2}{*}{3B} & Caption Only & 7.3 & 9.3 & 71.9 & 58.3 & 59.9 & 52.8 & 67.0 \\ 
 &  & Caption+VQA & 34.8 & 39.3 & 82.8 & 74.5 & 80.7 & 68.8 & {\bf 84.0} \\ 
 \cmidrule{2-10}
 & \multirow{2}{*}{11B} & Caption Only & 11.0 & 15.0 & 68.1 & 55.1 & 66.5 & 56.4 & 72.0 \\ 
 & & Caption+VQA & {\bf 46.0} & {\bf 47.9} & {\bf 85.3} &	{\bf 77.0} &	{\bf 85.0} &	{\bf 71.2}  &	{\bf 84.0} \\ 
\bottomrule[1.5pt]
\end{NiceTabular}
}
\label{tab:training_ablation}
\end{table*}

{\bf VQAScore is effective with simple question-answers.} \autoref{tab:prompting} shows that VQAScore consistently performs well across various question templates. Notably, on the challenging Winoground and EqBen benchmarks, simple yet clear questions tend to yield the best results for all VQA models. Interestingly, \autoref{tab:prompting_answer} shows that computing the negative answer likelihood (e.g., --P(``No'')) often yields comparable results. Furthermore, concise answers like P(``Yes'') perform better than longer responses such as P(``Yes it does''). We believe that VQAScore's simplicity makes it a strong alternative to the widely adopted divide-and-conquer approaches~\cite{vpeval, tifa, davidsonian, t2vscore, huang2023t2i}, which depend on carefully crafted in-context prompts.

\begin{table*}[h]
  \caption{\textbf{Ablating question templates for VQAScore.} We ablate 16 question templates across the three VQA models on the challenging Winoground and EqBen benchmarks. We report the group score, where higher scores indicate better performance. We highlight that most questions yield comparable performance, with clearer questions (e.g., those ending with ``.. Please answer yes or no.'') outperforming more ambiguous ones like ``\{\}?''. We also note that CLIP-FlanT5 and InstructBLIP tend to be more stable across different question templates, while LLaVA-1.5 varies more.}
  \centering
\renewcommand{\arraystretch}{1.3}
\scalebox{0.54}{
\begin{NiceTabular}{lccccccc}
            \CodeBefore
            \Body
  \toprule[1.5pt]
  \multirow{2}{*}{\textbf{Question Template}} & \multicolumn{2}{c}{\textbf{CLIP-FlanT5}} &
  \multicolumn{2}{c}{\textbf{LLaVA-1.5}} & \multicolumn{2}{c}{\textbf{InstructBLIP}}
  \\ 
  \cmidrule(l){2-3} \cmidrule(l){4-5} \cmidrule(l){6-7}
  &    Winoground & EqBen & Winoground & EqBen & Winoground  & EqBen  \\ \midrule
  \multicolumn{4}{l}{\it Our default question} \\
  Does this figure show "\{\}"? Please answer yes or no.  & 46.0  & 47.9  & 29.8  & 35.0  & 28.5  & {\bf 38.6} \\
  \midrule
  \multicolumn{4}{l}{\it Paraphrased yes-or-no questions} \\
  Is this figure showing "\{\}"? Please answer yes or no.  & {\bf 46.5}  & 48.6  & 26.8  & 35.0  & 28.2  & 35.0 \\
  Does this photo show "\{\}"? Please answer yes or no.  & 44.0  & 49.3  & 30.5  & 31.4  & 28.7  & 33.6 \\
  Does this picture show "\{\}"? Please answer yes or no.  & 44.5  & 48.6  & 30.2  & 38.6  & 29.5  & 32.9 \\
  Does this image show "\{\}"? Please answer yes or no.  & 43.2  & 47.9  & 29.2  & 30.7  & 28.2  & 32.9 \\
  Does it show "\{\}"? Please answer yes or no.  & 43.8  & 49.3  & 24.5  & 28.6  & 28.2  & 35.7 \\
  Does "\{\}"? Please answer yes or no.  & 43.8  & 49.3  & 31.8  & 37.1  & 28.7  & 32.1 \\
  Is "\{\}" an accurate description of this figure? Please answer yes or no.  & 43.5  & 47.9  & 27.5  & 30.0  & 27.3  & {\bf 38.6} \\
  Can "\{\}" be seen in this figure? Please answer yes or no.  & 40.8  & 49.3  & 25.8  & 27.9  & 26.8  & 32.9 \\
  "\{\}"? Please answer yes or no.  & 44.8  & {\bf 52.1}  & 32.5  & 30.0  & {\bf 30.2}  & 35.7 \\
  \midrule
  \multicolumn{4}{l}{\it Other questions} \\
  "\{\}"?  & 41.0  & 47.9  & 24.0  & 19.3  & 25.8  & 27.1 \\
  Does this figure show "\{\}"?  & 44.8  & 49.3  & 25.8  & 27.1  & 27.5  & 37.1 \\
  Does this figure show "\{\}"? Answer the question using a single word or phrase.  & 44.8  & 47.1  & {\bf 35.0}  & 39.3  & 26.8  & 37.1 \\
  What is the answer to the following question? "Does this figure show "\{\}"?"  & 42.0  & 45.0  & 20.8  & 32.1  & 27.8  & 35.7 \\
  Based on the image, respond to this question with a short answer: "Does this figure show "\{\}"?"  & 42.5  & 45.7  & 33.2  & {\bf 42.9}  & 27.8  & 35.0 \\
  The question "Does this figure show "\{\}"?" can be answered using the image. A short answer is   & 42.8  & 46.4  & 18.2  & 31.4  & 27.3  & 36.4 \\
\bottomrule[1.5pt]
\end{NiceTabular}
}
\label{tab:prompting}
\end{table*}

\begin{table*}[h]
  \caption{\textbf{Ablating answer formats for VQAScore.} Our analysis of the Winoground and EqBen benchmarks shows that extracting the negative answer likelihood yields comparable results, e.g., P(``Yes'') performs similarly to the negation of P(``No''). Furthermore, concise answers are more effective than longer responses like ``Yes it does''.}
  \centering
\renewcommand{\arraystretch}{1.3}
\scalebox{0.6}{
\begin{NiceTabular}{llccccccc}
            \CodeBefore
            \Body
  \toprule[1.5pt]
  \multirow{2}{*}{\textbf{Question Template}} & \multirow{2}{*}{\textbf{Answer}} & \multicolumn{2}{c}{\textbf{CLIP-FlanT5}} &
  \multicolumn{2}{c}{\textbf{LLaVA-1.5}} & \multicolumn{2}{c}{\textbf{InstructBLIP}}
  \\ 
  \cmidrule(l){3-4} \cmidrule(l){5-6} \cmidrule(l){7-8}
  &  &  Winoground & EqBen & Winoground & EqBen & Winoground  & EqBen  \\ \midrule
  \multirow{2}{*}{Does this figure show "\{\}"? Please answer yes or no.} & P(Yes)  & 46.0  & {\bf 47.9}  & {\bf 29.8}  & 35.0  & {\bf 28.5}  & {\bf 38.6} \\
  & --P(No)  & {\bf 46.3} & {\bf 47.9} & 27.5 & {\bf 37.1} & 28.0 & 32.9 \\
  \midrule
  \multirow{2}{*}{Does this figure show "\{\}"? Please answer correct or wrong.} & P(Correct)  & 18.0  & 30.7  & 21.8  & 32.9  & 24.8  & 30.7 \\
   & --P(Wrong)  & 36.0 & 31.4 & 18.3 & 20.0 & {\bf 28.5} & 35.0 \\
  \midrule
  \multirow{2}{*}{Does this figure show "\{\}"? Please answer true or false.} & P(True)  & 29.8  & 39.3  & 31.0  & 34.3  & 25.8  & 32.9 \\
   & --P(False)  & 42.5 & 37.9 & 27.0 & 30.0 & {\bf 28.5} & 33.6 \\
  \midrule
  \multirow{2}{*}{Does this figure show "\{\}"?} & P(Yes it does)  & 17.0  & 25.7  & 15.5  & 22.9  & 17.8  & 25.7 \\
   & --P(No it does not)  & 30.3 & 23.6 & 16.8 & 30.7 & 23.0 & 22.9  \\

\bottomrule[1.5pt]
\end{NiceTabular}
}
\label{tab:prompting_answer}
\end{table*}

\section{Details of Baseline Methods}
\label{sec:method_details}
In this section, we detail the implementation of the baseline methods and explore the reasons behind their failures.

{\bf Metrics based on vision-language models (CLIPScore/BLIPv2Score).} To calculate CLIPScore, we use the same CLIP-L-336 model~\cite{clipscore} of CLIP-FlanT5 and LLaVA-1.5. For BLIPv2Score, we use the ITM (image-text-matching) head~\cite{blipv2} from the largest BLIPv2-ViT-G variant. For an in-depth analysis of how these discriminatively pre-trained VLMs behave as bags-of-words models, we refer readers to previous studies~\cite{lin2024revisiting, aro, kamath2023text, winoground}.

{\bf Metrics finetuned on human feedback (PickScore/ImageReward/HPSv2).} We use the official code and model checkpoints to calculate these metrics. Specifically, PickScore~\cite{pickscore} and HPSv2~\cite{hpsv2} finetune the CLIP-H model, and ImageReward~\cite{imagereward} finetunes the BLIPv2, using costly human feedback from either random web users or expert annotators. Our experiments on the Winoground and EqBen benchmarks (\autoref{tab:vqascore_results_winoground_eqben}) show that these metrics perform no better than random chance, likely because the discriminative pre-trained VLMs bottleneck their performance due to bags-of-words behaviors. In addition, their finetuning datasets may lack compositional texts. Finally, we observe that human annotations can be noisy or subjective, especially when these annotators are not well trained (e.g., random web users of the Pick-a-pic dataset~\cite{pickscore}). We discuss these issues in \appref{sec:eval_details}. We leave it to future work to finetune VQAScore with human feedback.

{\bf Visual programming methods (VisProg/ViperGPT/VPEval).} We follow the official implementation of these methods. For VisProg~\cite{visprog} and ViperGPT~\cite{viper}, we apply the same VQAScore prompt (``Does this figure show "\{text\}"? Please answer yes or no.''). However, these methods struggle with compositional texts, e.g., Winoground~\cite{winoground}. For instance, given the text ``{\tt someone talks on the phone happily while another person sits angrily}'', VisProg simply requests a yes-or-no answer from a VQA model, without decomposing. ViperGPT generates the below program that overlooks the action relation:
\begin{mdframed}[linewidth=1pt, linecolor=black, leftmargin=2cm, rightmargin=2cm, backgroundcolor=gray!20, roundcorner=5pt]
\scriptsize
\textcolor{blue}{\# Text is "someone talks on the phone happily while another person sits angrily"}  \\
\textcolor{blue}{\# Below is the incorrect program generated by ViperGPT that ignores action relation}  \\
def execute\_command(image) -> int: \\  
\hspace*{10mm}image\_patch = ImagePatch(image)\\    
\hspace*{10mm}person\_patches = image\_patch.find("person")\\    
\hspace*{10mm}if len(person\_patches) < 2: \\
\hspace*{20mm}return 0\\    \hspace*{10mm}person\_patches.sort(key=lambda x: x.horizontal\_center)\\    \hspace*{10mm}person1\_patch = person\_patches[0]\\    
\hspace*{10mm}person2\_patch = person\_patches[1]\\    
\hspace*{10mm}person1\_happy = person1\_patch.verify\_property("person", "happy")\\    
\hspace*{10mm}person2\_angry = person2\_patch.verify\_property("person", "angry")\\    
\hspace*{10mm}if person1\_happy and person2\_angry:\\        
\hspace*{20mm}return 1\\    
\hspace*{10mm}else:\\        
\hspace*{20mm}return 0
\end{mdframed}

\noindent For VPEval~\cite{vpeval}, we follow its ``open-ended evaluation program'' designed for compositional texts. Nonetheless, we observe that it occasionally generates erroneous or nonsensical programs, like asking a VQA model ``what is the person doing while talking on the phone?'' and expecting an answer of ``happily''.

{\bf Divide-and-conquer using VQA (TIFA/VQ2/Davidsonian).} We first note that divide-and-conquer methods are the most popular in recent text-to-visual evaluation~\cite{dalle3, huang2023t2i, t2vscore, dreamsync}. Therefore, we comprehensively analyze all open-source methods, ensuring fair comparison by using the same VQA models as for VQAScore. Specifically, \autoref{tab:vqa_based_methods_on_winoground_eqben} already shows that our simple VQAScore surpasses the more complex TIFA~\cite{tifa}, VQ2~\cite{vq2}, and Davidsonian~\cite{davidsonian} across all VQA models (e.g., InstructBLIP-FlanT5-11B, LLaVA-1.5-13B, CLIP-FlanT5-11B). TIFA uses a finetuned Llama-2 to generate multiple-choice QA pairs, returning the answer accuracy of a VQA model as the alignment score. Davidsonian uses a more sophisticated pipeline by prompting ChatGPT to generate yes-or-no QA pairs while avoiding inconsistent questions. For example, given the text ``the moon is over the cow'', if a VQA model already answers ``No'' to ``Is there a cow?'', it then skips the follow-up question ``Is the moon over the cow?''. VQ2~\cite{vq2} uses a finetuned FlanT5 to generate free-form QA pairs and computes the average score of P(answer | image, question). However, these methods often generate nonsensical QA pairs, as shown in \autoref{tab:divide_errors}. Lastly, \autoref{tab:vqa_based_methods_on_winoground_eqben_tifa} confirms that using (a) a single question template {\em without decomposition} and (b) the {\em likelihood} of ``Yes'' is much more effective than decomposition using Davidsonian~\cite{davidsonian} or checking if the model can directly generate ``Yes''.

\begin{table*}[h]
\centering
\renewcommand{\arraystretch}{1.3}
\caption{\textbf{Ablation on question decomposition and answer generation versus likelihood.} For a fair comparison, we apply all methods to the same CLIP-FlanT5 model. Our end-to-end VQAScore (using the default question template) outperforms question decomposition using Davidsonian~\cite{davidsonian} or direct answer generation (i.e., checking if the generated answer is ``Yes'').
}
\scalebox{0.7}{
\begin{NiceTabular}{lllccccccc}
            \CodeBefore
            \Body
\toprule[1.5pt]
\multirow{2}{*}{\textbf{VQA Model}} & 
\multirow{2}{*}{\textbf{Question Template(s)}} & 
\multirow{2}{*}{\textbf{Scoring}} & \multicolumn{3}{c}{\textbf{Winoground}} &
\multicolumn{3}{c}{\textbf{EqBen}}
\\ 
\cmidrule(l){4-6} \cmidrule(l){7-9}
  &  & &    Text & Image & Group & Text & Image  & Group \\ \midrule

\multirow{4}{*}{CLIP-FlanT5-11B} & \multirow{2}{*}{Davidsonian~\cite{davidsonian}} & Generation & 16.3 & 11.5 & 9.8 & 17.1 & 11.4 & 11.4 \\
&  & VQAScore & 41.0 & 38.3 & 28.3 &	45.7 &   47.9 &     35.0 \\
\cmidrule(l){2-9}
& \multirow{2}{*}{\footnotesize Does this figure show "\{\}"? Please answer yes or no.} & Generation & 15.3 & 15.3 &	 15.3 &	21.4 & 21.4 & 21.4 \\
&  & VQAScore & {\bf 60.0} &	{\bf 57.5} &	{\bf 46.0} &	{\bf 59.3} &	{\bf 63.6} &	{\bf 47.9}\\

\bottomrule[1.5pt]
\end{NiceTabular}
}
\label{tab:vqa_based_methods_on_winoground_eqben_tifa}
\end{table*}

\begin{table*}[h]
\centering
\caption{\textbf{Failure cases of divide-and-conquer methods (TIFA, VQ2, and Davidsonian).} We show generated question-and-answer pairs of TIFA, VQ2, and Davidsonian on three Winoground texts. These methods often generate irrelevant or erroneous QA pairs (highlighted in \textcolor{red}{red}), especially with more compositional texts.}
\scalebox{0.75}{
\begin{NiceTabular}{M{0.15\linewidth} M{0.6\linewidth} M{0.4\linewidth}}
\CodeBefore
    \Body
\toprule[1.2pt]
 \textbf{Method} & \textbf{Generated questions} & \textbf{Candidate answers (correct answer choice in bold)} \\ \midrule

\multicolumn{3}{c}{Text: ``{\tt the moon is over the cow}''}\\
\multirow{2}{*}{TIFA} & Is the moon over the cow? & \textbf{yes}, no \\
& Is the moon over or under the cow? & \textbf{over}, under, next to, behind \\ \cmidrule{1-3}
\multirow{2}{*}{VQ2} & \textcolor{red}{What part of the sun is above the cow?} & {\bf the moon} \\
  & \textcolor{red}{What is the name of the moon over the cow?} & {\bf the moon} \\ \cmidrule{1-3}
\multirow{3}{*}{Davidsonian} & Is there a moon? & \textbf{yes}, no \\
  & Is there a cow? & \textbf{yes}, no \\ 
  & Is the moon over the cow? & \textbf{yes}, no \\ 
  \toprule[1.2pt]

\multicolumn{3}{c}{Text: ``{\tt someone talks on the phone happily while another person sits angrily}''}\\
\multirow{2}{*}{TIFA} & \textcolor{red}{Who is talking on the phone?} & \textbf{someone}, no one, everyone, someone else \\
& \textcolor{red}{Who is sitting angrily?} & \textbf{person}, animal, robot, alien \\ \cmidrule{1-3}
\multirow{2}{*}{VQ2} & \textcolor{red}{Who has a good time on the phone?} & {\bf someone} \\
  & \textcolor{red}{What part of the life does someone talk to?} & {\bf the phone} \\ \cmidrule{1-3}
\multirow{3}{*}{Davidsonian} & \textcolor{red}{Is the someone happy?} & \textbf{yes}, no \\
  & \textcolor{red}{Is there another person?} & \textbf{yes}, no \\ 
  & Is there a phone? & \textbf{yes}, no \\ 
  \toprule[1.2pt]

  \multicolumn{3}{c}{Text: ``{\tt all paper airplanes fly on a curved path except for one which takes a straight one}''}\\
\multirow{2}{*}{TIFA} & \textcolor{red}{Are the paper airplanes flying on a curved path?} & \textbf{yes}, no \\
& \textcolor{red}{Are the paper airplanes flying on a curved path or a straight path?} & \textbf{curved path}, straight path, wavy path, zigzag path \\ \cmidrule{1-3}
\multirow{2}{*}{VQ2} & \textcolor{red}{What type of airplanes fly on a straight path?} & {\bf all paper airplanes} \\
  & \textcolor{red}{All paper airplanes fly on what?} & {\bf a straight path} \\ \cmidrule{1-3}
\multirow{3}{*}{Davidsonian} & \textcolor{red}{Do paper airplanes fly on a curved path?} & \textbf{yes}, no \\
  & \textcolor{red}{Is there one paper airplane?} & \textbf{yes}, no \\ 
  & Do paper airplanes fly? & \textbf{yes}, no \\ 

\bottomrule[1.2pt]
\end{NiceTabular}
}
\label{tab:divide_errors}
\end{table*}

{\bf GPT4-Vision-based methods (GPT4-Eval/VIEScore).} We follow the official prompts from GPT4-Eval~\cite{zhang2023gpt} and VIEScore~\cite{viescore} to ask GPT4-Vision~\cite{gpt4} to directly generate an alignment score (in text format) for an image-text pair (e.g., 0 to 100). For detailed prompts, we direct readers to the respective papers or codebases. Note that we cannot use GPT4-Vision for VQAScore because its API currently does not expose likelihoods of generated answers. Nonetheless, we posit that using VQAScore on stronger VQA models like GPT4-Vision can outperform text-based alignment score generation as done by ~\cite{viescore, zhang2023gpt}.

{\bf T2VScore-A(lignment).} T2VScore-A~\cite{t2vscore} is a divide-and-conquer method specifically designed for video-text alignment. When reporting T2VScore-A~\cite{t2vscore} (based on GPT4-Vision), we calculate the pairwise accuracy~\cite{deutsch2023ties} using scores released by the authors. However, the authors do not provide the corresponding T2VScore-A outputs for other VQA models (e.g., InstructBLIP).

\section{Details of Alignment Benchmarks}
\label{sec:eval_details}

In this section, we provide details on evaluation metrics and benchmarks in the main paper.

{\bf (Meta-)evaluation metrics for human agreement (Pairwise accuracy/Pearson/Kendall).} To meta-evaluate metrics (e.g., VQAScore) on benchmarks that provide 1-5 Likert scale ratings (e.g., TIFA160~\cite{tifa}), we primarily report the pairwise accuracy (with tie calibration) as advocated by Deutsch et al.~\cite{deutsch2023ties}. Pairwise accuracy effectively addresses ties common in human ratings, unlike the classic Kendall metric which ignores ties. We direct readers to \cite{deutsch2023ties} for detailed equations and provide a brief overview below. For a dataset containing $M$ image-text pairs, there are two score vectors of size $M$ each: one for human ratings and one for metric scores. \cite{deutsch2023ties} evaluates pairwise rankings to determine if human and metric scores agree, i.e., if one image-text pair scores higher, lower, or ties with another image-text pair across both human and metric scores. Additionally, \cite{deutsch2023ties} performs tie calibration to optimize for the best tie threshold in metric scores. We emphasize that Pairwise accuracy (with tie calibration) is more reliable and interpretable. Unlike the Pearson coefficient, \cite{deutsch2023ties} does {\em not} assume linear correspondence between human ratings and metric scores. Furthermore, when compared to the Kendall coefficient (which also measures correct pairwise ranking decisions), \cite{deutsch2023ties} provides an accuracy value ranging from 0 to 1, making it easier to interpret. For completeness, \autoref{tab:tifa_all_results} and \autoref{tab:flickr8k_all_results} report all three metrics on TIFA160~\cite{tifa} and Flickr8K~\cite{clipscore}.

{\bf TIFA160~\cite{tifa}.} TIFA160 collects 160 text prompts from four sources: MSCOCO captions~\cite{coco}, DrawBench~\cite{imagen}, PartiPrompts~\cite{parti}, and PaintSkill~\cite{dalleval}. Each text prompt is paired with five text-to-image models, generating a total of 800 image-text pairs. Furthermore, Davidsonian~\cite{davidsonian} labels these image-text pairs using 1-5 Likert scale for human evaluation. \autoref{tab:tifa_all_results} shows that our VQAScore consistently surpasses prior methods across all three meta-evaluation metrics.

\begin{table}[h!]
\centering
\caption{{\bf Evaluating agreement with human judgment on text-to-image benchmark TIFA160~\cite{tifa, davidsonian}.} We report Pairwise accuracy, Pearson, and Kendall(-b), with higher scores indicating stronger agreement between human and metric scores. VQAScore based on our CLIP-FlanT5 consistently surpasses all other methods.
}
\scalebox{0.83}{
\begin{NiceTabular}
{lccc}
        \CodeBefore
            \rectanglecolor{softgreen}{17-1}{17-4}
            \rectanglecolor{softgreen}{22-1}{22-4}
            \rectanglecolor{softgreen}{27-1}{27-4}
        \Body
\toprule[1.5pt]
\multirow{2}{*}{\textbf{Method}} & \multirow{2}{*}{\textbf{\small \begin{tabular}{@{}c@{}}Pairwise \\ Acc~\cite{deutsch2023ties}\end{tabular}}} & \multicolumn{2}{c}{\textbf{\small \textcolor{gray}{Old metrics}}}  \\ \cmidrule(r){3-4} 
& & \textcolor{gray}{\small Pearson} & \textcolor{gray}{\small Kendall} \\
\midrule
\multicolumn{4}{l}{\textit{Baselines}} \\
CLIPScore~\cite{clipscore} & 55.8 & \textcolor{gray}{29.6} & \textcolor{gray}{19.9} \\
BLIPv2Score~\cite{blipv2} & 57.5 & \textcolor{gray}{35.6} & \textcolor{gray}{23.3}   \\
\midrule
\multicolumn{4}{l}{\textit{HumanFeedback-based}} \\
ImageReward~\cite{imagereward} & 67.3 & \textcolor{gray}{61.5} & \textcolor{gray}{43.8}  \\ 
PickScore~\cite{pickscore} &  59.4 & \textcolor{gray}{39.8} & \textcolor{gray}{27.4}  \\ 
HPSv2~\cite{hpsv2} & 55.2 & \textcolor{gray}{30.1} & \textcolor{gray}{19.1}   \\
\midrule
\multicolumn{4}{l}{\textit{GPT4-Vision-based}} \\
GPT4V-Eval~\cite{zhang2023gpt} & 64.0 & \textcolor{gray}{58.9} & \textcolor{gray}{46.8}   \\
VIEScore~\cite{viescore} & 63.9 & \textcolor{gray}{61.2} & \textcolor{gray}{47.4}  \\ 
\midrule
\multicolumn{4}{l}{\textit{InstructBLIP-based}} \\
TIFA~\cite{tifa} & 60.0 & \textcolor{gray}{56.5} & \textcolor{gray}{44.0}  \\ %
VQ2~\cite{vq2} & 50.8 & \textcolor{gray}{12.1} & \textcolor{gray}{9.4}\\ %
Davidsonian~\cite{davidsonian} & 61.8 &  \textcolor{gray}{{63.4}} & \textcolor{gray}{48.5}   \\ %
\textbf{VQAScore (Ours)} & {70.1} & \textcolor{gray}{58.5} & \textcolor{gray}{{49.7}}  \\ \midrule
{\textit{LLaVA-1.5-based}} \\
TIFA~\cite{tifa} & 60.4 & \textcolor{gray}{49.3} & \textcolor{gray}{38.1}  \\
VQ2~\cite{vq2} & 48.7 & \textcolor{gray}{4.7} & \textcolor{gray}{5.1}   \\
Davidsonian~\cite{davidsonian} & 54.3 & \textcolor{gray}{55.6} & \textcolor{gray}{{45.4}}   \\
\textbf{VQAScore (Ours)} & 66.4 & \textcolor{gray}{{58.9}} & \textcolor{gray}{41.9}   \\ \midrule 
{\textbf{CLIP-FlanT5-based (Ours)}} \\
TIFA~\cite{tifa}  & 60.4 & \textcolor{gray}{46.3} & \textcolor{gray}{36.0}   \\
VQ2~\cite{vq2}  & 49.0 & \textcolor{gray}{3.9} & \textcolor{gray}{5.6}  \\
Davidsonian~\cite{davidsonian} & 61.4 & \textcolor{gray}{49.0} & \textcolor{gray}{37.0}   \\
\textbf{VQAScore (Ours)} & {\bf 71.2} & \textcolor{gray}{{\bf 66.2}} & \textcolor{gray}{{\bf 51.9}}   \\ 
\bottomrule[1.5pt]
\end{NiceTabular}
}
\label{tab:tifa_all_results}
\end{table}

{\bf Flickr8K~\cite{clipscore}.} We report on the image-to-text evaluation benchmark Flickr8K-CF to show that VQAScore can evaluate image captions in a {\em reference-free} manner like CLIPScore~\cite{clipscore} (without using reference captions of each image). Specifically, Flickr8K-CF contains 145K binary quality judgments collected via CrowdFlower for 48K (image, caption) pairs. Each pair receives at least 3 binary judgments, with human ratings calculated as the mean proportion of ``yes'' annotations for each pair. \autoref{tab:flickr8k_all_results} demonstrates that our VQAScore outperforms all prior art, including reference-based metrics such as BLEU-4, CIDEr, and RefCLIPScore~\cite{clipscore}.

\begin{table*}[h!]
\centering
\renewcommand{\arraystretch}{1.3}
\caption{{\bf Evaluating agreement on image-to-text benchmark Flickr8K~\cite{clipscore}.} We report Pairwise accuracy, Pearson, and Kendall, with higher scores indicating better performance for all metrics. In this benchmark, each image-caption pair is rated by at least three annotators. VQAScore achieves superior performance compared to existing methods like RefCLIPScore and CIDEr in a reference-free manner (without using the reference captions of the images as provided by the dataset).}
\scalebox{0.8}{
\centering
\begin{NiceTabular}{llccc}
    \CodeBefore
      \rectanglecolor{softgreen}{13-1}{16-5}
    \Body
\toprule[1.5pt]
\multirow{2}{*}{\textbf{Method}} & \multirow{2}{*}{\textbf{Model}} & \multirow{2}{*}{\textbf{\small \begin{tabular}{@{}c@{}}Pairwise \\  Acc~\cite{deutsch2023ties}\end{tabular}}} & \multicolumn{2}{c}{\textbf{\small \textcolor{gray}{Old metrics}}}  \\ \cmidrule(r){4-5} 
& & & \textcolor{gray}{\small Pearson} & \textcolor{gray}{\small Kendall} \\
\midrule
\multicolumn{3}{l}{\textit{Reference-based metrics}} \\
BLEU-4 & - & 78.1 & \textcolor{gray}{19.8} & \textcolor{gray}{16.9} \\
METEOR & - & 78.4 & \textcolor{gray}{36.8} & \textcolor{gray}{22.3} \\
ROUGE & - & 78.0 & \textcolor{gray}{32.6} & \textcolor{gray}{19.9} \\
CIDEr & - & 79.3 & \textcolor{gray}{46.1} & \textcolor{gray}{24.6} \\
SPICE & - & 78.2 & \textcolor{gray}{35.7} & \textcolor{gray}{24.4} \\
RefCLIPScore~\cite{clipscore} & ViT-B/32 & 78.2 & \textcolor{gray}{47.9} & \textcolor{gray}{36.4} \\
\midrule
{\textit{Reference-free metrics using CLIPScore}} \\
\multirow{2}{*}{CLIPScore~\cite{clipscore}} & ViT-B/32 & 77.8 & \textcolor{gray}{44.4} & \textcolor{gray}{34.4}  \\
 & ViT-L/14-336px & 78.2 & \textcolor{gray}{46.5} & \textcolor{gray}{34.7}  \\
 \midrule
\midrule
{\textit{Reference-free metrics using VQAScore}} \\
\multirow{3}{*}{\textbf{VQAScore (Ours)}} & InstructBLIP & 81.5 & \textcolor{gray}{58.2} & \textcolor{gray}{36.0} \\
 & LLaVA-1.5 & 82.4 & \textcolor{gray}{61.9} & \textcolor{gray}{36.4} \\
 & CLIP-FlanT5 (Ours) & {\bf 83.1} & \textcolor{gray}{{\bf 65.4}} & \textcolor{gray}{{\bf 36.7}} \\
\bottomrule[1.5pt]
\end{NiceTabular}
}
\label{tab:flickr8k_all_results}
\end{table*}

{\bf EvalCrafter~\cite{evalcrafter, t2vscore}.} We use the text-to-video evaluation benchmark EvalCrafter with 1-5 Likert scales collected by T2VScore~\cite{t2vscore} for assessing video-text alignment. This benchmark contains 700 prompts paired with five text-to-video models such as Pika~\cite{pikalab}, Gen2~\cite{gen2}, and Floor33~\cite{floor33}. By default, we average the VQAScore of all 36 frames from the 3-second videos. \autoref{tab:t2vscore_frame_ablation} also shows that sampling as few as four frames can achieve near-optimal performance.

\begin{table}[h!]
\centering
\caption{{\bf Ablating the number of sampled frames for the text-to-video benchmark EvalCrafter~\cite{t2vscore}.} We report the pairwise accuracy~\cite{deutsch2023ties} of VQAScore for one, four, and all (36) uniformly sampled frames. VQAScore achieves the best performance with 36 frames and near-optimal performance with as few as four frames.}
\scalebox{0.83}{
\begin{tabular}{lccc}
\toprule[1.2pt]
\multirow{2}{*}{\textbf{Model}} & \multicolumn{3}{c}{\textbf{\small 
Sampled Frames}}  \\ \cmidrule(r){2-4} 
& {\small One} & {\small Four} & {\small All} \\
\midrule
InstructBLIP & 65.4 & 65.8 & 65.7  \\ %
LLaVA-1.5 & 63.2 & 63.7 & 63.6   \\
\midrule
CLIP-FlanT5 & {\bf 65.8} &  {\bf 66.5} &  {\bf 66.5}    \\
\bottomrule[1.2pt]
\end{tabular}
}
\label{tab:t2vscore_frame_ablation}
\end{table}

{\bf StanfordT23D~\cite{wu2024gpt}.} We use the text-to-3D evaluation benchmark StanfordT23D and collect our own 1-5 Likert scales for assessing 3D-text alignment. We follow the same annotation procedure as GenAI-Bench (\autoref{sec:genai_bench_details}) and gather 3 human ratings per 3D-text pair, spanning six text-to-3D models (Latent-Nerf~\cite{latentnerf}/Magic-3D~\cite{magic3d}/MVDream~\cite{mvdream}/DreamFusion~\cite{dreamfusion}/Instant3D~\cite{instant3d}/Shap-E~\cite{shape}) across 60 prompts. For human annotators, we provide a 3x3 grid view of each 3D asset, with 9 views sampled uniformly across camera angles. By default, we average the VQAScore of all 120 provided views. However, \autoref{tab:stanfordt23d_views_ablation} shows that using the same 3x3 grid view (that requires only a single pass) can achieve near-optimal performance.

\begin{table}[h!]
\centering
\caption{{\bf Ablating the number of sampled views and input formats for text-to-3D benchmark StanfordT23D~\cite{wu2024gpt}.} We report the pairwise accuracy~\cite{deutsch2023ties} with higher scores indicating better performance. Interestingly, using a single grid layout (2x2 or 3x3) image often performs almost as well as averaging VQAScores across 4 or 9 views.  
}
\scalebox{0.83}{
\begin{tabular}{lccccc}
\toprule[1.2pt]
\multirow{2}{*}{\textbf{Model}} & \multicolumn{5}{c}{\textbf{\small 
Sampled Views}}  \\ \cmidrule(r){2-6}  
 & {\small Uniform (4)} & {\small Grid (2x2)} & {\small Uniform (9)}  & {\small Grid (3x3)} & {\small All (120)}   \\
\midrule
InstructBLIP & 67.4 & 67.4 & 68.0 & 68.1 &  68.1 \\ %
LLaVA-1.5 & 64.5 & 64.8 & 64.9  & 64.9 & 64.9  \\
CLIP-FlanT5 & {\bf 68.1}  & {\bf 67.8} & {\bf 68.5} & {\bf 68.4} &  {\bf 68.6}   \\
\bottomrule[1.2pt]
\end{tabular}
}
\label{tab:stanfordt23d_views_ablation}
\end{table}

{\bf Pic-a-pick~\cite{pickscore}.} We find that the text-to-image evaluation benchmark, Pic-a-pick, contains an excessive amount of NSFW (sexual/violent) content and incorrect labels, likely due to an inadequate automatic filtering procedure. Specifically, after manually reviewing the test set of 500 samples, we find that 10\% contain inappropriate content (e.g., ``{\em zentai}'' and ``{\em Emma Frost as an alluring college professor wearing a low neckline top}'') and approximately 50\% had incorrect labels. This may also account for the inferior performance of PickScore. As a result, we manually filter the test set to obtain a clean subset of 100 prompts paired with 200 images for evaluating binary accuracy. We also remove all tied labels due to their subjective nature. We will release this subset of Pick-a-pic for reproducibility.

{\bf SeeTrue~\cite{vq2} (DrawBench/EditBench/COCO-T2I).} We utilize the binary match-or-not labels collected by SeeTrue~\cite{vq2} for the three benchmarks. These benchmarks consist of individual image-text pairs, where some pairs are correctly paired and others are not. We follow their original evaluation protocols to report the AUROC (Area Under the Receiver Operating Characteristic curve), taking into account all possible classification thresholds.

{\bf Winoground~\cite{winoground} and EqBen~\cite{eqben}.} In our study, we use the entire Winoground dataset consisting of 400 pairs of image-text pairs. For EqBen, because the official test set includes low-quality images (e.g., very dark or blurry pictures), we analyze the higher-quality EqBen-Mini subset of 280 pairs of image-text pairs, as recommended by their official codebase. These two benchmarks evaluate image-text alignment via matching tasks: each sample becomes 2 image-to-text matching tasks with one image and two candidate captions, and 2 text-to-image matching tasks with one caption and two candidate images. The text (and image) score is awarded 1 point only if {\em both} matching tasks are correct. The final group score is awarded 1 point only if {\em all} 4 matching tasks are correct. Importantly, we discover that these benchmarks (especially Winoground) test advanced compositional reasoning skills crucial for understanding real-world prompts, such as counting, comparison, differentiation, and logical reasoning. These advanced compositions operate on basic visual entities, which themselves can be compositions of objects, attributes, and relations.

\end{document}